\documentclass{article} 
\usepackage{iclr2026_conference,times}

\usepackage{amsmath,amsfonts,bm}

\def\eqref#1{equation~\ref{#1}}

\def\1{\bm{1}}

\DeclareMathAlphabet{\mathsfit}{\encodingdefault}{\sfdefault}{m}{sl}
\SetMathAlphabet{\mathsfit}{bold}{\encodingdefault}{\sfdefault}{bx}{n}

\PassOptionsToPackage{hyphens}{url} 

\usepackage[utf8]{inputenc}
\usepackage[table,xcdraw]{xcolor}
\usepackage{makecell}
\usepackage{booktabs}
\usepackage{float}
\usepackage{graphicx}
\usepackage{tabularx,array}
\usepackage{longtable}
\usepackage{lscape}
\usepackage{multirow}
\usepackage{algorithm}
\usepackage{algpseudocode}
\usepackage{enumitem}
\usepackage{afterpage}
\usepackage{tcolorbox}
\usepackage{listings}
\usepackage{url}
\usepackage{hyperref}

\newcommand{\meanstd}[2]{\ensuremath{#1{\scriptstyle\,\pm\,#2}}}

\title{CP-Agent: Context‑Aware Multimodal Reasoning for Cellular Morphological Profiling under Chemical Perturbations}

\author{
Yuxin Zhang\textsuperscript{1,*},
Yiyao Li\textsuperscript{2,*},
Ping Shu Ho\textsuperscript{4},
Simon See\textsuperscript{4},
Zhenqin Wu\textsuperscript{2,\dag},
Kevin Tsia\textsuperscript{1,3,5,\dag}
\\[0.5em]
\textsuperscript{1}Department of Electrical and Computer Engineering, The University of Hong Kong \\
\textsuperscript{2}School of Computing and Data Science, The University of Hong Kong\\
\textsuperscript{3}School of Biomedical Engineering, The University of Hong Kong\\
\textsuperscript{4}Nvidia AI Technology Center\\
\textsuperscript{5}Advanced Biomedical Instrumentation Centre
\\[0.5em]}

 \iclrfinalcopy
\begin{document}


\maketitle
\begingroup
\renewcommand\thefootnote{}
\footnotetext{\textsuperscript{*}Equal contribution. \textsuperscript{\dag}Corresponding authors. Project page: https://github.com/letitia-zhang/CP-Agent}
\endgroup

\vspace{-1em}
\begin{abstract}

Cell Painting combines multiplexed fluorescent staining, high‑content imaging, and quantitative analysis to generate high-dimensional phenotypic readouts to support diverse downstream tasks such as mechanism-of-action (MoA) inference, toxicity prediction, and construction of drug–disease atlases. However, existing workflows are slow, costly and difficult to interpret. Approaches for drug screening modeling predominantly focus on molecular representation learning, while neglecting actual experimental context (e.g., cell line, dosing schedule, etc.), limiting generalization and MoA resolution. We introduce CP-Agent, an agentic multimodal large language model (MLLM) capable of generating mechanism-relevant, human-interpretable rationales for cell morphological changes under drug perturbations. At its core, CP-Agent leverages a context-aware alignment module, CP-CLIP, that jointly embeds high-content images and experimental metadata to enable robust treatment and MoA discrimination (achieving a maximum F1-score of 0.896). By integrating CP-CLIP outputs with agentic tool usage and reasoning, CP‑Agent compiles rationales into a structured report to guide experimental design and hypothesis refinement. These capabilities highlight CP-Agent’s potential to accelerate drug discovery by enabling more interpretable, scalable, and context-aware phenotypic screening -- streamlining iterative cycles of hypothesis generation in drug discovery.

\end{abstract}

\section{Introduction}

High‑content imaging with Cell Painting has become a workhorse for scalable phenotypic drug discovery. This technique, integrating advanced microscopy, multiplexed fluorescent staining and quantitative image analysis, allows us to establish high-dimensional morphological cell profiles that capture rich multiscale cellular responses to chemical perturbations. These profiles have been proven valuable in supporting mechanism-of-action (MoA) inference~\citep{tian2023combining}, toxicity prediction~\citep{ewald2025cell}, hit triage~\citep{vincent2020hit}, and drug repurposing~\citep{fredin2024cell}, while also enabling the construction of reference atlases and improved target deconvolution~\citep{moffat2017opportunities}.

In Cell Painting workflows, cells are perturbed under diverse conditions and the experimental context is not a nuisance to control but a signal to model. For instance, dose and time define trajectories; cellular background modulates pathway readouts (Appendix~\ref{cell_painting_experiment}). The resulting profiles guide follow‑up experiments and can advance phenotype-driven drug discovery. However, Cell Painting-based drug discovery remains limited by several challenges: (i) complex intermediate dependencies: Morphological responses are highly context-dependent. For example, concentration-dependent profiles show low correlations across dose levels (Pearson r = 0.21-0.26)~\citep{trapotsi2022cell}, and MoA prediction is sensitive to cell line context~\citep{seal2024decade}. Ignoring these structures conflates biology with acquisition artifacts and wastes the valuable metadata; (ii) convergent morphologies: Compounds with distinct mechanisms may induce morphological readouts convergence, reducing MoA resolution, thereby complicating the extraction of standardized, interpretable descriptors. (iii) Lack of semantic grounding: Representing image embeddings as unstructured feature vectors restricts their capacity for semantic reasoning and downstream biological inference.


Recently, various AI methods have been introduced to Cell Painting datasets, such as generative approaches to synthesize images under perturbations~\citep{navidi2024morphodiff,cross2023class,palma2025predicting}, multimodal frameworks integrating chemical and genetic annotations with cell painting images~\citep{sanchez2023cloome} ~\citep{fradkin2024molecules, lu2025cellclip}. For example, CLOOME firstly introduced a CLIP-style model to align Cell Painting images with molecular structures. MolPhenix and CellCLIP further extend this direction by leveraging strong unimodal foundation models to align the molecule. However, many existing models offer visual embeddings as black-box features, which lack semantic interpretability. Moreover, experimental context is often under‑used: metadata is appended via late fusion or treated as unstructured text, yielding less informative representations and hindering iterative, closed‑loop experimental design. Meanwhile, emerging multimodal large language models (MLLMs) offer reasoning capabilities and have been applied in diverse biological domains, such as genomics, biomedical imaging, and omics data analysis~\citep{zhang2024mm,lin2025bridging,liu2024geneverse,hu2024advancing,zhang2024multimodal}. Yet their applications in drug screening remain underexplored.

In this work, we introduce CP-Agent, a context-aware, agentic MLLM framework for Cell Painting drug perturbation screening. At its core is CP‑CLIP, a contrastive alignment module that jointly embeds Cell Painting images and structured experimental context, including drug compounds and other essential experimental conditions, enhancing the biological relevance of cell morphology. The model is pretrained on 1.9 million image-context pairs, with a customized token injection strategy that embeds key fields for better alignment. Comprehensive evaluations across curated classification tasks show that CP-CLIP outperforms general-purpose baselines. Built on this perception layer, CP-Agent integrates tool-augmented reasoning and task-adapted MLLMs grounded in phenotype descriptors and MoA ontologies to generate structured, interpretable outputs. Together, this agentic system supports scalable and interoperable phenotypic analysis, enabling cross-study generalization and providing actionable insights for assay prioritization and iteration, thereby accelerating hypothesis generation and improving decision-making in phenotypic drug discovery.

\section{Method}
\vspace{-0.5em}
\subsection{Dataset}
\label{dataset}
We employed three open-access Cell Painting datasets, consisting of approximately 1.9 million pairs: BBBC021~\citep{caie2010high}, CPJUMP1~\citep{chandrasekaran2024three}, and RxRx3~\citep{fay2023rxrx3}, encompassing diverse compound-induced phenotypes. Each image-context pair comprises a microscopy image and its associated experimental context (e.g., cell lines, experimental treatment conditions) We curated compounds to ensure traceable MoA labels across datasets. For each collection, we matched SMILES representations of the perturbing chemical compounds to ChEMBL, retrieved their targets and MoAs, and retained only compounds with publicly resolvable MoA names. A summary of the curated multi-dataset setting is provided in Table~\ref{dataset-table}. More details about dataset backgrounds are provided in Appendix~\ref{apx:datasets}.
\begin{table}[h]
\caption{Summary of datasets used in this study}
\label{dataset-table}
\centering
\small
\renewcommand{\arraystretch}{1.3}
\setlength{\tabcolsep}{4pt}
\begin{tabular}{@{}lllllll@{}}
\toprule
\textbf{Dataset} &
\textbf{Cell line} &
\textbf{Channel} &
\textbf{Compound} &
\textbf{Concentration} &
\textbf{Time} &
\textbf{Image Pair} \\
\midrule
BBBC021 & MCF-7 (p53 WT) & 3 & 34 & Variable 8-point half-log & 24 h & 144,411\\
CPJUMP1 & U2OS, A549 & 5 & 62 & 5.0~\textmu{}M & 24 h, 48 h & 562,687\\
RXRX3 & HUVEC & 6 & 380 & Fixed 8-point half-log & $\sim$20~h & 1,265,984 \\
\bottomrule
\end{tabular}
\end{table}

The training set comprises 1,846,436 image–text pairs, while the validation set contains 9,395 pairs. For zero-shot evaluation, we curated a held-out set of compounds spanning all three datasets, selected to assess generalization to unseen perturbations. 

\subsection{Molecular Drug Encoding}
\label{Molecular_Encoding}
Several established approaches map compound perturbations to vector representations, enabling alignment with image embeddings and facilitating multimodal learning~\citep{winter2019learning, wu2025chemberta}. For instance, SMILES-based (e.g., ChemBERTa) and graph-based models learn molecular embeddings from structure, often using RDKit for preprocessing. Alternatively, one can compute continuous molecular descriptor embeddings (e.g., physicochemical and topological descriptors), formalized as a parameterized feature extractor: $\phi_{\mathrm{desc}}(x ; P)=\left[f_1\left(x ; P_1\right), f_2\left(x ;P_2\right), \ldots, f_d\left(x ; P_d\right)\right] \in \mathbb{R}^d$, where $x$ is an input molecular representation (e.g., SMILES strings or molecular graphs), and each $f_i(x; P_i)$ extracts a specific property, forming a $d$-dimensional real-valued feature vector. In contrast, binary fingerprint embeddings that encode the presence/absence of substructures (e.g., Morgan/circular, MACCS, or path-based fingerprints)~\citep{bento2020open} $\phi_{\mathrm{fp}}: \mathcal{M} \rightarrow\{0,1\}^d \text { or } \mathbb{N}_0^d$, yield binary or count-based encoding over the molecular space $\mathcal{M}$.

\subsection{CP-CLIP: Reprocessing}
To harmonize~\textbf{Cell Painting images} across datasets with varying resolution and signal quality, we defined a channel-wise preprocessing step: $\mathcal{P}: \mathbb{R}^{H_0 \times W_0} \rightarrow \mathbb{R}^{H \times W}$, applied independently to each fluorescence channel. This includes Contrast Limited Adaptive Histogram Equalization (CLAHE), random Laplacian sharpening, and gamma correction, yielding enhanced images $\tilde{I}=\mathcal{P}(I)$. Enhanced single-channel images are then cropped into $512\times512$ patches and stacked, yielding input tiles $x_p \in \mathbb{R}^{512 \times 512 \times C}$. For each perturbation tile $x_p$, a corresponding control tile $x_c \in \mathbb{R}^{512 \times 512 \times C}$ is independently sampled from a matching control set $\Omega\left(x_p\right)$, which share all experimental contexts (e.g., plate, cell line, channel) with $x_p$, except for the perturbation compound. That is $x_c \sim \mathcal{U}\left(\Omega\left(x_p\right)\right)$. The final image branch input is formed by concatenating the grayscale perturbation and control tiles along the channel dimension, $\hat{x}=\operatorname{concat}\left(x_p, x_c\right) \in \mathbb{R}^{512 \times 512 \times 2}$. This paired design encourages the model to learn the contrasts between treated and untreated states.

\textbf{Molecular descriptors} are projected via a fixed dimensional mapping $f_{\text {desc }}:\mathcal{X} \rightarrow \mathbb{R}^d$, where each feature dimension corresponds to a predefined physicochemical or topological property (See Appendix~\ref{apx:RDKit2D}). Let $v=f_{\operatorname{desc}}(x) \in \mathbb{R}^d$ denote the raw descriptor vector for compound $x \in \mathcal{X}$. To ensure numerical stability and comparability across compounds, dimensions containing undefined values (e.g., NaNs or Infs) are removed, and z-score normalization is applied independently to each feature dimension $\tilde{v}_i=\frac{v_i-\mu_i}{\sigma_i}$.

To account for the compound-specific dosing scheme, each molecule is represented by a normalized dosing pair$\left[\rho_{\max }, s(C)\right]$, 
where $\rho_{\max }$ denotes the molecular mass-normalized maximum concentration (in $mg/mL$), and $s(C)$ is the log-scaled dose step index corresponding to a given concentration. Let $M \in \mathbb{R}_{>0}$ denote the molecular weight (in $Da$ or $g/mol$), and $C_{\max } \in \mathbb{R}_{>0}$ the nominal maximum concentration (in $\mu$M). So, the molecular maximum mass concentration is given by:
\begin{equation}
    \rho_{\max }[\mathrm{mg} / \mathrm{mL}]:=\frac{M[\mathrm{Da}] \cdot C_{\max }[\mu \mathrm{M}]}{10^6}
\end{equation}
where the denominator $10^6$ reflects the conversion from $\mu$M and $Da$ to $\mathrm{mg} / \mathrm{mL}$. While for each titration point $C \in\left\{C_1, \ldots, C_8\right\}$, a pseudo-step index is computed on a log scale to reflect dilution ratios:
\begin{equation}
  s(C):=\frac{\log _{10}\left(C_{\max }\right)-\log _{10}(C)}{\Delta \log }, \quad \Delta \log =0.5
\end{equation}
where the denominator 0.5 corresponds to the log-fold change between adjacent titration levels in a 2-fold serial dilution protocol. A detailed derivation is provided in Appendix~\ref{apx:log_dose_Indexing}.

For \textbf{observation time}, let $t \in \mathbb{R}_{\geq 0}$ denote time in days. Temporal normalization rescales $t$ into the unit interval via:
$
    \tilde{t}=\frac{t}{T_{\max }}, \quad \text { with } T_{\max }=112
$. The 112-day (16-week) window reflects the FDA's stopping rule, adopted by~\citet{watkins2022economic} in their pharmacoeconomic analysis. These representations ensure that the input space remains consistent across compounds with varying dosing schemes and time-points. 

\subsection{CP-CLIP: Context-Aware Token Projection}
\label{Metadata-Aware_Projection}
\begin{figure}[t]
\begin{center}
\includegraphics[width=0.9 \linewidth]{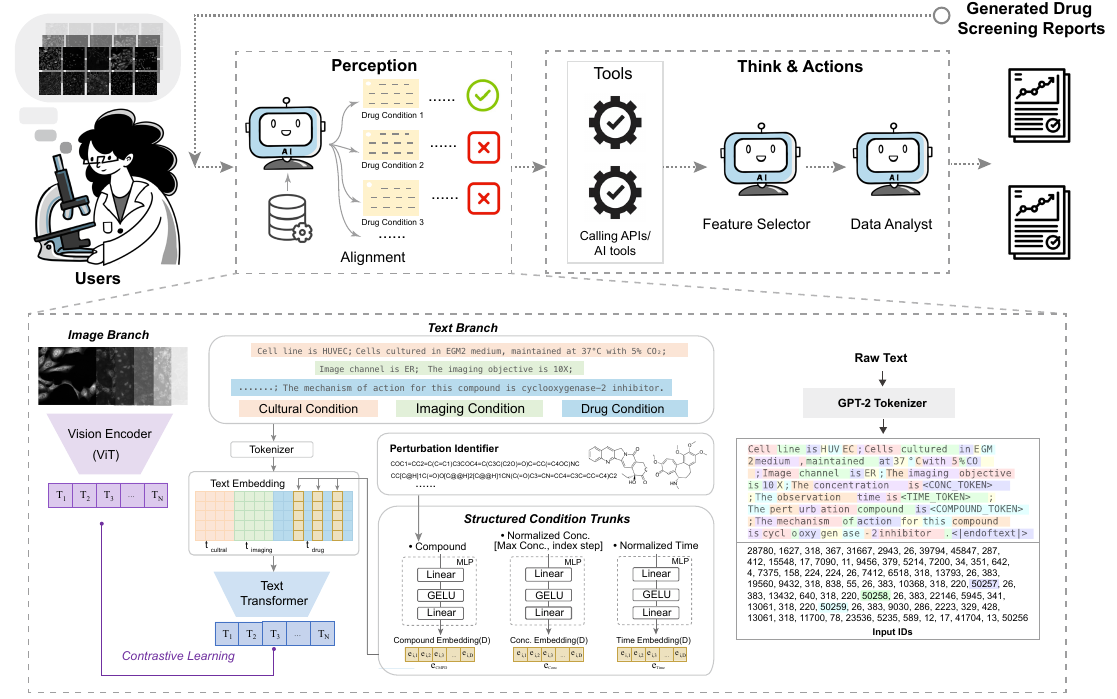}
\end{center}
\caption{Illustration of the CP‑agent (top) and CP‑CLIP (bottom). CP-Agent connects perception, memory retrieval, and modular analysis into a unified pipeline for generating reports for Cell Painting experiments. CP-CLIP forms the backbone of the CP-Agent’s perception module, providing joint embeddings of Cell Painting images and structured experimental context.}\label{fig: general_workflow}
\end{figure}
Our contrastive framework uses a structured text encoder tailored to the metadata obtained from drug screening experiments (Figure~\ref{fig: general_workflow}, bottom). Each experiment is represented as a prompt-like sequence composed of cell culture, imaging, and drug compound perturbation conditions. So the “raw text” refers to structured experimental metadata such as cell line, culture medium, imaging parameters, compound identity, dosage, time and other cultural information if have. These contextual descriptions are first composed into a natural language-style sentence and tokenized into input IDs using the standard GPT-2. To accommodate structured context and consistent representations of perturbing compounds, we introduced field-specific placeholder tokens (i.e. \texttt{<CMPD>}, \texttt{<CONC>}, \texttt{<TIME>}) for compound descriptors $z_{\text {cmpd }}=\phi_{\text {desc }}(x ; P) \in \mathbb{R}^d$, normalized concentration $z_{\text {conc }}=\left[\rho_{\max }, s(C)\right] \in \mathbb{R}^2$, and normalized time $z_{\text {time }}=\tilde{t} \in \mathbb{R}$. The special placeholder tokens are directly inserted into the text sequence and registered into the tokenizer’s vocabulary. During tokenization, they are automatically recognized as atomic units and their positions are preserved without being split or altered. Their embeddings are then dynamically computed via field-specific Multilayer Perceptron (MLP) trunks $f_*: \mathbb{R}^{d^{\prime}} \rightarrow \mathbb{R}^D$:
\begin{equation}
    \begin{aligned}
    e_{\text {cmpd }} & =f_{\text {cmpd }}\left(z_{\text {cmpd}}\right) \in \mathbb{R}^D \\
    e_{\text {conc }} & =f_{\text {conc }}\left(z_{\text {conc}}\right) \in \mathbb{R}^D \\
    e_{\text {time }} & =f_{\text {time }}\left(z_{\text {time}}\right) \in \mathbb{R}^D
    \end{aligned}
\end{equation}
where $f_{\text{cmpd}}, f_{\text{conc}}, \text{and } f_{\text{time}}$ are lightweight MLP trunks encoding compound identity, concentration, and time-point used in place of the placeholders. The resulting text input is a hybrid sequence:
\begin{equation}
    X=[\mathrm{CLS}, t_1, t_2, \ldots, \underbrace{e_{\text {cmpd }}}_{<\mathrm{CMPD}>}, \ldots, \underbrace{e_{\text {conc }}}_{<\text {CONC}>}, \ldots, \underbrace{e_{\text {time }}}_{<\text {TIME}>}, \ldots]
\end{equation}
This hybrid sequence, combining standard subword embeddings $t_i \in \mathbb{R}^D$ with structured embeddings $\mathbf{e}_* \in \mathbb{R}^D$ from field-specific MLPs, is fed into the text Transformer to produce final text representation. Implementation details are in Appendix~\ref{token_replace_modules}. By replacing placeholder tokens with learned embeddings, the model fuses continuous metadata with discrete language tokens in a shared embedding space. The text encoder thus captures both experimental signals and linguistic coherence, enabling better semantic alignment. 

\begin{figure}[t]
\label{workflow}
\begin{center}
\includegraphics[width=0.95 \linewidth]{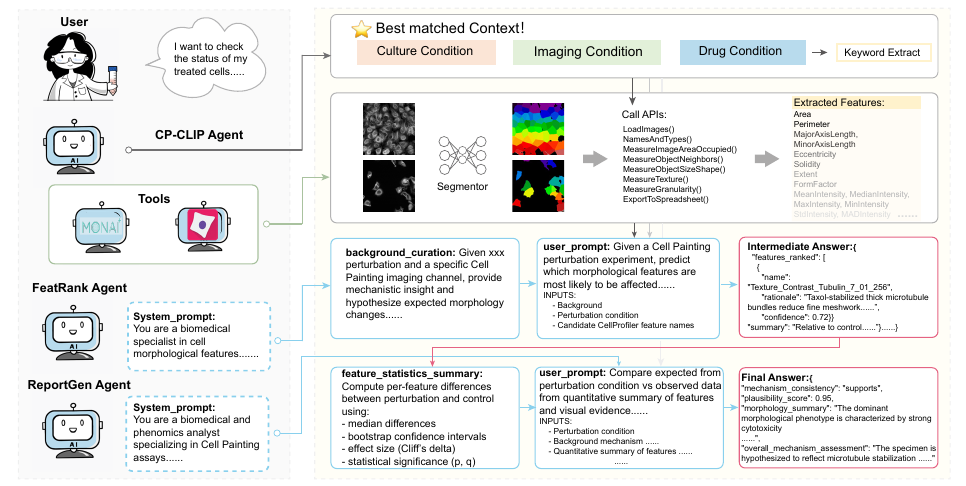}
\end{center}
\caption{Automated cell-phenotype assessment pipeline of CP-Agent. Upon user query, CP-CLIP retrieves the relevant experimental context to guide cell segmentation and feature extraction. Downstream agents then rank morphological changes and generate interpretable, end-to-end phenotype reports.}\label{agent_workflow}
\end{figure}

\subsection{CP-Agent workflow}
CP-Agent adopts a modular, memory-augmented architecture that connects perception, tooling, and analysis into a single-pass pipeline (Figure~\ref{fig: general_workflow}, top). Given user-provided Cell Painting images, a lightweight memory retriever powered by CP-CLIP fetches the most probable experimental context (i.e., cell line, fluorescence channels, imaging settings, chemical perturbations). Once the experimental context is retrieved, the pipeline proceeds to visual analysis. Rather than relying on vision backbones that produce holistic, biologically opaque embeddings, we extract handcrafted single-cell morphological features. These interpretable representations are processed by a modular, MLLM-driven agent architecture, where the MLLM serves as a policy layer that dynamically routes tasks to interchangeable tools and integrates their outputs.
We frame this system as ``agentic'' in the sense of \textit{procedural autonomy}~\citep{xu2025comprehensive}: unlike reinforcement learning-based planners, CP-Agent employs the MLLM as a cognitive controller within a structured workflow. It relies on the model's learned reasoning capabilities—rather than fixed logical scripts—to dynamically prioritize morphological features, interpret statistical distribution shifts, and synthesize mechanism-level hypotheses based on retrieved experimental context.

We instantiate this concept on fluorescence Cell Painting data via a specialized CP-Agent workflow (Figure~\ref{agent_workflow}), which comprises the following steps:


\begin{itemize}[left=0pt,labelsep=0.5em]
\item \textbf{CPContext Agent}  
Given paired Cell Painting images (control vs. perturbation) acquired under matched conditions, the \textit{CPContext Agent} employs a pre-trained CP-CLIP retriever to obtain experimental context from a curated knowledge base. Simultaneously, it harmonizes metadata via controlled-vocabulary tagging and channel labeling to generate standardized descriptors. Retrieved context is routed both (A) as a context bundle to \textit{FeatRank Agent}, \textit{ReportGen Agent}, and (B) as metadata keywords to the \textit{CellFeat Agent}.
\item \textbf{ChannelSeg Agent}
Given Cell Painting images, the \textit{ChannelSeg Agent} performs nuclei instance segmentation on DNA-stained channels and whole-cell segmentation on non-DNA channels (e.g., RNA, Actin, ER, etc.). It outputs channel-specific instance masks, which are passed to the \textit{CellFeat Agent}.
\item \textbf{CellFeat Agent} 
Given Cell Painting images, corresponding masks, and harmonized metadata, the \textit{CellFeat Agent} extracts per-cell morphological, intensity, texture, granularity, neighborhood, and occupancy features using a configured CellProfiler pipeline (Appendix~\ref{apx:CellProfiler}). Output is routed both (A) as extracted feature items to the \textit{FeatRank Agent} for mechanism-aware selection, and (B) as channel-wise single-cell feature matrices to the ~\textit{StatSynth Agent} for statistical evidence synthesis. 
\item \textbf{FeatRank Agent} 
Given extracted feature items and experimental context, the \textit{FeatRank Agent} scores and ranks features by their likelihood of being influenced by the perturbation. It generates confidence-weighted rationales to support prioritization. Output is routed as a prioritized feature list with explanations to the ~\textit{StatSynth Agent}.
\item \textbf{StatSynth Agent}
Given the prioritized feature list, full feature matrices, and experiment-level context, the \textit{StatSynth Agent} computes per-feature statistical evidence between control and perturbation conditions based on the prioritized features. It summarizes distribution shifts, effect sizes, confidence intervals, and statistical significance. Outputs are routed as statistical summaries and interpretations to the~\textit{ReportGen Agent} for final report composition.
\item \textbf{ReportGen Agent}
Given statistical summaries, prioritized features, visual exemplars, and experimental context, the \textit{ReportGen Agent} composes an integrated interpretation of the perturbation’s biological impact. It identifies key morphological shifts and evaluates their consistency with expected cellular responses to infer plausible mechanisms. The resulting report summarizes these findings, provides follow-up recommendations and visualizations, and is delivered to the users for downstream access.
\end{itemize}
The agent tool stack integrates both classical and learning-based components. For segmentation, we fine-tuned VISTA-2D~\citet{he7vista3d} for 20 epochs using diverse augmentation strategies to mitigate optics-induced batch effects. The model generates channel-specific masks that enable biologically consistent segmentation across diverse imaging conditions. More details regarding dataset preparation and training of the segmentation model are provided in Appendix~\ref{VISTA-2D}. The \textit{StatSynth Agent} is tasked with reasoning over high-dimensional single-cell morphological data (typically 30–300 cells per image), which is impractical for direct LLM application due to length constraints and noise~\citep{fang2024large}. Instead, we curate agentic tools that  (i) aggregate summary statistics for key features, and (ii) quantify distribution shifts between control and perturbed samples. These compact, interpretable summaries support reliable LLM-based  reasoning. Detailed procedures for this step are provided in Appendix~\ref{apx:statistics_synthesizer}.

\section{Experiments and Results}

\begin{table}[htbp]
\caption{Model performance on classification tasks}
\label{classification-table}
\centering
\setlength{\tabcolsep}{4pt} 
\renewcommand{\arraystretch}{1.2} 
\resizebox{\textwidth}{!}{%
\begin{tabular}{ccc|*{11}{c}} 
\toprule
\textbf{Model} & \textbf{Cell line} & \textbf{Channel} & \multicolumn{11}{c}{\textbf{Perturbation Compound}} \\
\cmidrule(lr){4-13}
& & & Flindokalner  & Racecadotril & AZM-475271 & Misoprostol & Trazodone & Orantinib & Rufinamide & Lumiracoxib & BIRB-796 & Methoxsalen & Macro-avg \\
\midrule
\rowcolor{gray!20}
Random Guessing & 0.25 & 0.143 & 0.10 & 0.10 & 0.10 & 0.10 & 0.10 & 0.10 & 0.10 & 0.10 & 0.10 & 0.10 & 0.10 \\
Grok-4 & 0.448  & 0.228  & 0.215  & 0.174  & 0.0 & 0.0  & 0.410  & 0.190  &  0.034 &  0.0 & 0.0  & 0.0 & 0.102 \\
GPT-5 &  0.377 &  0.439 & 0.059 &  0.168 & 0.0  & 0.0 & 0.353 & 0.0  & 0.0 &  0.0 &  0.0 &  0.0 & 0.074 \\
Claude-4-Sonnet & 0.450  & 0.198  & 0.0  &  0.00 & 0.0 & 0.057 & 0.0  & 0.0  & 0.0  & 0.0  & 0.211 &  0.0  & 0.027\\
Gemini-2.5-Pro &  0.526 &  0.628 & 0.0  &  0.0 & 0.0 & 0.0 & 0.0  & 0.023  & 0.0  & 0.0  & 0.045 &  0.0 & 0.007\\
\midrule
CLOOME ViT-B/16 & - & - & 0.784 & 0.784 & 0.729 & 0.854 & 0.623 & 0.849 & 0.653 & 0.619 & 0.854 & 0.800 & 0.755\\
CLIP ViT-B/16 & \textbf{1.000}  & 0.955  & 0.776  & 0.680  & 0.661  & 0.216  & 0.629 & 0.447  &  0.500&  0.600 &  0.575 &  0.642 & 0.657\\
SigLIP-ViT-B/16 & \textbf{1.000}  &  0.925  & 0.734  &   0.471 & 0.515   &  0.826  & 0.291  &  0.638  & 0.395  & 0.272  &  0.604  &  0.400 & 0.514 \\
\makecell[c]{CP-CLIP SigLIP-ViT-B/16\\\textit{(descriptor)}} & \textbf{1.000} & 0.934 &  0.685 & 0.442  &  0.485 & 0.776 & 0.351  &  0.860 & 0.255  & 0.186  & 0.660 & 0.620 & 0.532\\
\makecell[c]{CP-CLIP ViT-B/16\\\textit{(fingerprint)}} & \textbf{1.000}  & \textbf{0.991}  & 0.839  & 0.862 &  0.891 &  0.875 & \textbf{0.913}  & 0.914 &  0.894 & 0.840  & \textbf{0.971} & 0.875 & 0.887  \\
\makecell[c]{CP-CLIP ViT-B/16\\\textit{(descriptor)}} & \textbf{1.000} & 0.882 &  0.907  &  0.869  & 0.857  &  \textbf{0.942}   & 0.848  & \textbf{0.940} &  0.884  & \textbf{0.854} & 0.932 & 0.922 & \textbf{0.896}\\
\makecell[c]{CP-CLIP ViT-L/16\\\textit{(descriptor)}} & \textbf{1.000} & 0.849 & \textbf{0.928} &  \textbf{0.880}  &  \textbf{0.896}  & 0.846  &  0.843   & 0.929  & \textbf{0.911} &  0.819  & 0.915 & \textbf{0.941}  & 0.891\\
\bottomrule
\end{tabular}
}
\end{table}
\begin{table}[htbp]
\caption{Unseen drugs similarity score}
\label{unseen-drug}
\centering
\setlength{\tabcolsep}{4pt}
\renewcommand{\arraystretch}{1.2}
\resizebox{\textwidth}{!}{%
\begin{tabular}{c c c c c c c c c c}
\toprule
Model  & Regorafenib & Sacubitril & Buparlisib & Dexamethasone & Nimodipine & AZ258 & Nilotinib & MG-132 & Average\\
\midrule
CLIP ViT-B/16  & $\meanstd{0.207}{0.082}$ & $\meanstd{0.2058}{0.104}$ & $\meanstd{0.289}{0.046}$ & $\meanstd{0.3601}{0.049}$ & $\meanstd{0.377}{0.039}$ & $\meanstd{0.328}{0.069}$ & $\meanstd{0.174}{0.080}$ & $\meanstd{0.346}{0.072}$ & $0.286$\\
SigLIP ViT-B/16  & $\meanstd{0.038}{0.082}$  & $\meanstd{0.095}{0.099}$  & $\meanstd{0.129}{0.073}$  & $\meanstd{0.146}{0.091}$  & $\meanstd{0.183}{0.067}$  & $\meanstd{0.090}{0.186}$  & $\meanstd{-0.055}{0.103}$  & $\meanstd{0.143}{0.101}$& $0.096$  \\
\makecell[c]{CP-CLIP SigLIP-ViT-B/16\\\textit{(descriptor)}} & $\meanstd{0.378}{0.077}$  & $\meanstd{0.420}{0.193}$  & $\meanstd{0.323}{0.102}$  & $\meanstd{0.503}{0.130}$  & $\meanstd{0.515}{0.075}$  & $\boldsymbol{\meanstd{0.488}{0.115}}$  & $\meanstd{0.303}{0.090}$  & $\meanstd{0.380}{0.114}$& $0.414$\\
\makecell[c]{CP-CLIP ViT-B/16\\\textit{(fingerprint)}}  & $\meanstd{0.297}{0.093}$ & $\meanstd{0.222}{0.072}$ & $\meanstd{0.375}{0.053}$ & $\meanstd{0.468}{0.052}$ & $\meanstd{0.461}{0.046}$ & $\meanstd{0.429}{0.120}$ & $\meanstd{0.210}{0.109}$ & $\meanstd{0.420}{0.081}$ & $0.360$ \\
\makecell[c]{CP-CLIP ViT-B/16\\\textit{(descriptor)}}  & $\meanstd{0.432}{0.098}$ & $\meanstd{0.412}{0.094}$ & $\meanstd{0.396}{0.043}$ & $\meanstd{0.503}{0.073}$ & $\meanstd{0.469}{0.032}$ & $\meanstd{0.468}{0.104}$ & $\boldsymbol{\meanstd{0.324}{0.085}}$ & $\meanstd{0.448}{0.081}$ & $0.432$\\
\makecell[c]{CP-CLIP ViT-L/16\\\textit{(descriptor)}}  & $\boldsymbol{\meanstd{0.455}{0.115}}$ & $\boldsymbol{\meanstd{0.445}{0.135}}$ & $\boldsymbol{\meanstd{0.408}{0.053}}$ & $\boldsymbol{\meanstd{0.530}{0.072}}$ & $\boldsymbol{\meanstd{0.523}{0.032}}$ & $\meanstd{0.448}{0.106}$ & $\meanstd{0.295}{0.089}$ & $\boldsymbol{\meanstd{0.448}{0.077}}$ & $\boldsymbol{0.444}$\\
\bottomrule
\end{tabular}
}
\end{table}


To assess the effectiveness of CP-Agent, we isolated and evaluated its core components before measuring end-to-end reporting quality: (a) CP-CLIP (context-aware retrieval and alignment): we evaluate its accuracy on in-distribution classification (seen-drug) and generalization (unseen-drug matching), ablations against MLLM baselines and CLIP variants; (b) Vision embedding structure: we evaluate whether CP-CLIP embeddings encode chemically grounded, dose- and MoA-dependent morphology; (c) Statistical synthesis and reporting: whether compact summaries enable robust comparisons between control and perturbation in the generated report. Finally, we assessed the effectiveness of full CP-Agent reports via expert review.


\subsection{Model Variants and MLLM Baselines}

\label{model_variants}

To contextualize the performance of our proposed model, we compared it against several leading MLLMs. Specifically, we included Grok-4~\citep{xai2025grok4}, GPT-5~\citep{openai2025gpt5}, Claude-4-Sonnet~\citep{anthropic2025claude4}, and Gemini-2.5-Pro~\citep{deepmind2025gemini25}, which have demonstrated strong performance across a range of general-purpose multimodal benchmarks. Following recent benchmarking protocols for MLLMs in biomedical and healthcare settings \citep{lozano2024mu, burgess2025microvqa}, we adopt a zero-shot, two-stage prompting pipeline: first, the models were prompted to curate background knowledge relevant to Cell Painting experiments; then, they were asked to answer multiple-choice questions about experimental conditions given the curated background knowledge, paired control and perturbation images, and masked textual prompts. To further narrow the adaptation gap and make the comparison more conservative, we additionally evaluated a few-shot variant in which the MLLMs were provided with a small visual memory bank (two labeled exemplars per class) before answering the same tasks. Detailed prompt templates and the corresponding zero-shot and few-shot results are reported in the Appendix~\ref{apx:MLLM}.

Alongside these MLLMs, we benchmarked multiple variants of our contrastive learning framework, CP-CLIP, which extends the CLIP architecture by integrating structured experimental context into training. As a baseline, we used the original CLIP model based on the ViT-B/16 vision backbone, retrained on natural language text aligned with Cell Painting images. All CP-CLIP variants enhance this setup by injecting serialized numerical metadata, as detailed in Section~\ref{Metadata-Aware_Projection}. We evaluated CP-CLIP variants that differ in compound encoding and loss function (See Appendix~\ref{apx:losses}), including: (i) a descriptor-based model used continuous molecular descriptors, and (ii) a fingerprint-based model used binary fingerprints. We also tested a SigLIP variant that uses a sigmoid-based pairwise contrastive objective~\citep{zhai2023sigmoid}. To assess the impact of vision model capacity on performance, we tested a CP-CLIP variant with ViT-L/16 vision backbone.

\begin{figure}[h]
\begin{center}
\includegraphics[width=0.95 \linewidth]{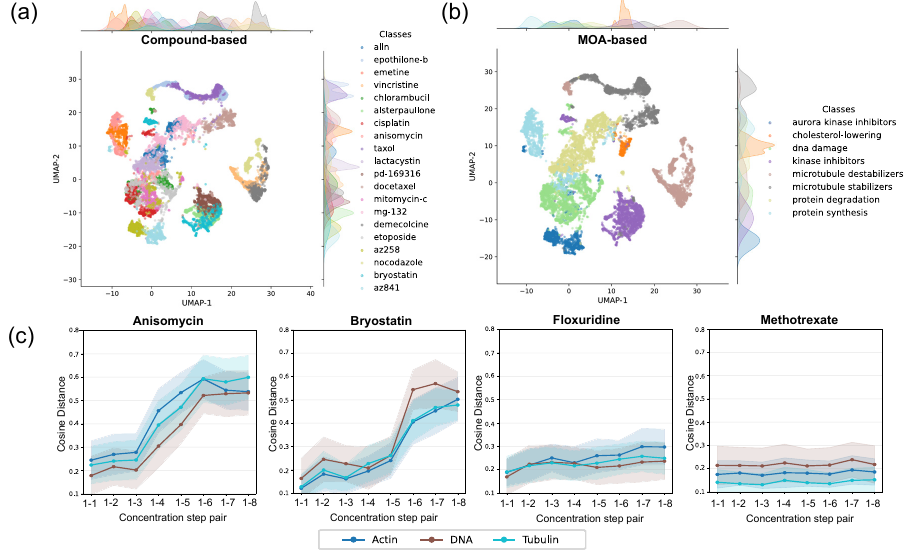}
\end{center}
\caption{CP-CLIP captures pharmacologically meaningful morphology. UMAP projections of CP-CLIP image embeddings, colored by (a) compound identity and (b) mechanism of action (MoA). The clear clustering indicates that the learned representation encodes biologically relevant morphology. (c) Concentration-dependent morphological changes are captured using image embeddings extracted from samples treated with varying compound doses.}\label{umaps}
\end{figure}

\subsection{Task I: seen-drug classification}

\label{seen-drug}
To benchmark in-distribution performance, we designed classification tasks across three categories: cell line, fluorescence channel and compound. The classification is performed retrieval-based inference by ranking cosine similarity scores between image embeddings and a set of candidate textual prompts, following the standard CLIP paradigm. Those contextual metadata includes both textual and numerical variables, which are encoded jointly as a natural language sequence, enabling prompt-based querying without the needs for task-specific heads. For example, for compound classification, 10 compounds were randomly sampled to form a balanced 10-class setting. Table~\ref{classification-table} summarizes the results. Among all general-purpose MLLMs, Gemini-2.5-Pro achieved the best performance on the cell line and channel prediction tasks (F1: 0.526 and 0.628). However, on compound classification, performance dropped sharply: All models fell below random baseline, except for Grok-4, which slightly exceeded it. Confusion matrices (Appendix~\ref{apx: MLLM_performance}) revealed near-zero F1 scores, indicating systematic failure in identifying perturbing chemical compounds and limited generalization of current MLLMs. In contrast, CP-CLIP consistently outperformed both the baseline CLIP and all MLLMs across tasks. Descriptor-based models slightly outperformed fingerprint-based ones on compound classification (F1: 0.891 vs. 0.887), indicating that continuous encodings provide richer chemical contexts. Scaling the vision encoder from ViT-B/16 to ViT-L/16 yielded no significant gain (F1: 0.896 vs. 0.891), indicating that a lightweight backbone suffices when paired with strong chemical priors. Taken together, these MLLMs results also constitute a “no-CPContext” baseline,  reinforcing the conclusion that without explicit perturbation-aware grounding, current MLLMs fail to extract meaningful biological signals from Cell Painting image. This emphasizes the essential role of CP-CLIP as the perception in CP-agent.

\subsection{Task II: unseen-drug matching}

To evaluate generalization, we performed zero-shot prompt–image matching on held-out compounds by computing cosine similarity between image and prompt embeddings (Table~\ref{unseen-drug}). The baseline CLIP model (ViT-B/16) yielded low alignment on unseen drugs (avg. similarity = 0.286), while CP-CLIP (descriptor, ViT-B/16) achieved 0.432, a 14.6\% absolute increase. Descriptor-based models also outperformed fingerprint-based ones (0.432 vs. 0.360), indicating that continuous encodings capture more relevant chemical contexts. Scaling the vision encoder from ViT-B/16 to ViT-L/16 further improved performance to 0.444, suggesting enhanced robustness to morphological variation. To provide a comparative reference, we also evaluated similarity on seen drugs (Appendix~\ref{tab:seen-drug}). Notably, performance on unseen drugs remained close, indicating strong generalization. Specifically, descriptor-based ViT-B/16 and ViT-L/16 models achieved 0.549/0.432 and 0.561/0.444 on seen/unseen drugs, suggesting that CP-CLIP captures mechanism-relevant biology, rather than memorizing labels. This zero-shot capability supports practical applications such as MoA hypothesis generation, hit prioritization, and generalization to novel perturbation contexts.

\subsection{vision embedding analyses}

\label{vision_embedding}
Figure~\ref{umaps}a-b shows UMAP projections of embeddings from CP-CLIP ViT-B/16 (descriptor). The UMAP projection reveals clustering by MoA, indicating the learned representation encodes pharmacologically meaningful morphology beyond compound identity.
Figure~\ref{umaps}c shows concentration-related patterns for four drugs selected from the BBBC021 and RxRx3 datasets. CP-CLIP embeddings exhibited clear dose–response trajectories, reflecting concentration-dependent morphological change. In particular, the sharp dose-responses observed for Anisomycin and Bryostatin are consistent with previous reports ~\citet{cranston1982further, marshall2002phase}. In contrast, drugs with minimal morphological impacts show flatter trends across dosage. More examples and a detailed explanation of this schematic are provided in the Appendix~\ref{Dose_response}.

\subsection{CP-Agent reports}
\label{sec:CP-Agent reports}

\label{reports}
\begin{figure}[t]
\begin{center}
\includegraphics[width=0.95 \linewidth]{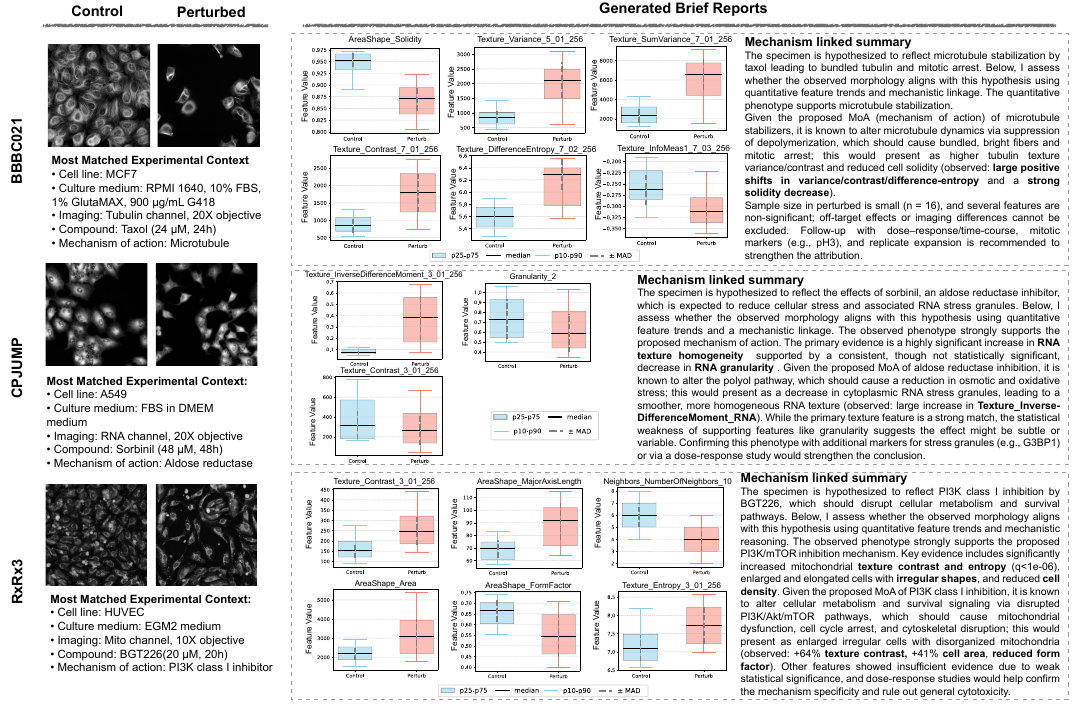}
\end{center}
\caption{Summary reports generated from CP-Agent. The examples show CP-Agent’s ability to recognize clear (Taxol), subtle (Sorbinil), and complex (BGT226) morphological responses, linking them to plausible biological mechanisms.}\label{fig:report_example}
\end{figure}
We present three case studies from different datasets to demonstrate CP-Agent generated reports (Figure~\ref{fig:report_example}): (i)\textit{Example 1 (BBBC021, MCF7 + Taxol)}: Taxol induces a clear~\textit{cytoskeletal phenotype} by stabilizing microtubules and arresting mitosis~\citep{kiwanuka2022effect}. CP-Agent detected the localized changes in tubulin texture and correctly linked them to microtubule stabilization and mitotic arrest, demonstrating its ability to recognize canonical, visually prominent phenotypes. (ii) \textit{Example 2 (CPJUMP, A549 + Sorbinil)}: Sorbinil is an aldose reductase inhibitor that produces a \textit{subtle and uncertain phenotype}~\citep{zietek2025cell}. CP-Agent detected modest shifts (e.g., smoother RNA texture, reduced granularity), and suggested potential stress granule suppression. Meanwhile, it also flagged ambiguity and suggested further validation, illustrating its ability to reason under uncertainty. (iii) \textit{Example 3 (RxRx3, HUVEC + BGT226)}: BGT226 is a PI3K/mTOR inhibitor, leading to a \textit{multi-compartment phenotype} affecting organelles, cell shape, and density~\citep{kampa2013cell}. By integrating mitochondrial texture, cell area, and density changes, CP-Agent inferred PI3K/mTOR inhibition, showcasing its capacity to synthesize complex morphological cues into mechanistic insights. Together, these cases show that CP-Agent adapts to diverse biological contexts, ranging from clear to ambiguous phenotypes, and generates biologically grounded summaries. Additional examples and reasoning details are provided in Appendix~\ref{additional_reports}.

We conducted an expert survey to assess whether LLM-based CP-Agent produces accurate and well-reasoned screening reports. Four LLMs (mentioned in Section~\ref{model_variants}) each generated reports for ten control–perturbation image pairs. Experts (N = 11), ranging from PhD students to professors in pharmacology or related fields, rated 40 reports (10 pairs × 4 models) on a 1–7 scale across ten criteria from~\citet{waqas2025reasoning}, covering language quality and reasoning quality. Full criteria definitions and examples are provided in Appendix~\ref{reasoning_criteria}. As shown in Figure~\ref{Appendix_reasoning}, most metrics received high scores across models. CP-Agent powered by GPT-5 showed the strongest overall reasoning performance, followed closely by Gemini-2.5-Pro. To further assess the interpretability and consistency of the CP-Agent framework, we conducted a systematic evaluation of the two LLM-powered modules: FeatRank Agent and ReportGen Agent. As shown in Table~\ref{FeatRank Agent's Repeatability}, the selected morphological features remained highly consistent across runs, indicating robust and stable feature prioritization. Table~\ref{tab:different_temperature} also revealed a stable corpus-level consistency of the report generated from ReportGen Agent.

\section{Discussion and Conclusion}
We present CP-Agent, a context-aware multimodal reasoning framework for interpretable analysis of Cell Painting drug responses. Its core, CP-CLIP, aligns imaging data with experimental context, enhanced by numerically grounded token injection. This yields strong generalization and outperforms baselines on multiple classification tasks. CP-Agent separates and coordinates perception, retrieval, analysis, and reporting into specialized agents (i.e.,CPContext, ChannelSeg, CellFeat, FeatRank, StatSynth, ReportGen). This enables an evidence-first workflow where CP-Agent converts high-dimensional morphological features, together with the experimental context, into compact, calibrated summaries that an MLLM synthesizes into interpretable narratives. Hence, CP-Agent allows end-to-end biological interpretability. Users can trace predicted mechanisms back to corresponding morphological features—from images to masks, features, statistics, and final explanations. Unlike histology tasks, where many agent-based pipelines can perform well without training by using a well-designed chain-of-thought with off-the-shelf MLLMs, our results show that zero-shot prompting for Cell Painting datasets consistently underperforms, and biologically grounded supervision is essential for meaningful reasoning. CP-Agent also generalizes to various imaging modalities such as quantitative phase imaging (QPI), digital holographic microscopy, and brightfield time-lapse imaging~\citep{lo2024information,siu2023optofluidic,zhang2023morphological,lee2025high} and integrates flexibly with tools like ilastik, Fiji, and Icy. Overall, it establishes a new paradigm for combining MLLMs with mechanistically grounded analysis, offering a foundation for next-generation AI systems in phenotypic drug discovery. Looking forward, the modular agentic architecture of CP-Agent could flexibly be extended for experimental planning (e.g., dose strategy refinement), multi-omics fusion, as well as causal priors for counterfactual reasoning. 
\subsubsection*{Ethical Statement}
This work does not involve human subjects, animal experiments, or personally identifiable data. All experiments are conducted on publicly available Cell Painting datasets. 

\subsubsection*{Reproducibility Statement}
All code, training scripts, and instructions necessary to reproduce our results are available at the anonymized repository:
https://github.com/letitia-zhang/CP-Agent

\subsubsection*{Acknowledgements}
The work is supported by Advanced Biomedical Instrumentation Center, the Research Grants Council (grant no. 17125121, 14125924, 17128225, RFS2021-7S06), the Innovation and Technology Commission of the Hong Kong Special Administrative Region of China (grant no. ITS/318/22FP, ITS/408/23FP).

\bibliography{iclr2026_conference}
\bibliographystyle{iclr2026_conference}

\appendix
\raggedbottom 
\section{Use of Large Language Models (LLMs)}
We used large language models (e.g., GPT-4) for non-substantive assistance during manuscript preparation. Specifically, LLMs were used to improve writing clarity, grammar, and phrasing, but not for generating scientific content or experimental design. All technical contributions, experiments, and interpretations were conceived and conducted by the authors. 

The authors take full responsibility for the content of the manuscript, including any text generatedor polished by the LLM. We have ensured that the [LM-generated text adheres to ethical guidelinesand does not contribute to plagiarism or scientific misconduct.
\section{Preliminaries and Background}
\subsection{High-content imaging}
\label{HCI}
High-content imaging (HCI) leverages automated microscopy and quantitative morphology to profile compound effects. Cell Painting stains multiple cellular components and extracts hundreds of single-cell features, producing high-dimensional representations that enable cross-perturbation comparisons, including compound clustering, target and pathway inference, and prediction of unannotated mechanisms~\citep{bray2016cell,odje2024unleashing}.

\subsection{Multidimensional Experimental Design in Cell Painting Assays}
\vspace{-0.5em}
\label{cell_painting_experiment}
Drug screening with Cell Painting involves diverse experimental factors that strongly shape cell morphology. (Overview of high-content imaging (HCI) can be referred to Appendix~\ref{HCI}). Key sources of variability include the cell line~\citep{LEJAL2025117530}, culture medium~\citet{harkness2019media}, incubation environment, and drug administration, each capable of inducing substantial morphological shifts. Drug libraries typically contain hundreds of thousands of molecules~\citep{huggins2011rational, liu2025impact}, with concentrations sampled using half-logarithmic dilution series to capture dose–response characteristics across orders of magnitude~\citep{choy2021high,miyajima2025parallel}. Meanwhile, temporal variables staging further increase complexity, as different observation time points can capture different call phases of treatment response, revealing both immediate and progressive morphological changes~\citep{beesabathuni2025image,LEJAL2025117530}. The interplay of experimental variables defines a high-dimensional space which condition combinations yield diverse morphological phenotypes.

\subsection{MLLM Agents for Bioinformatics}
Large language models (LLMs) are demonstrating growing potential across diverse domains of bioinformatics, with applications ranging from gene expression analysis~\citep{liu2024genotex} and drug discovery~\citep{averly2025liddia} to pathology image interpretation~\citep{lu2024multimodal}, spatial transcriptomics~\citep{wang2024jure}, and gene perturbation studies. Because datasets in these fields are often high-dimensional, recent efforts have increasingly turned to multimodal large language models (MLLMs), which integrate visual features from images with prior textual knowledge. Leveraging logical inference strategies such as deduction, induction, abduction, and analogy, MLLMs can support existing pipelines and facilitate novel scientific insights.

More recently, an emerging paradigm has focused on deploying MLLMs as autonomous or semi-autonomous agents to execute complex bioinformatics workflows~\citep{yiyao2025omicsnavigator, su2025biomaster}. Such agents integrate heterogeneous tools and interact through natural language, enabling biological data analysis guided by human instructions. While early studies highlight the promise of MLLM-driven agents in augmenting traditional pipelines, their scope has largely been limited to direct perception and recognition tasks. They remain insufficient for deeper understanding of complex biological processes and for generating novel hypotheses. Addressing this gap, we introduce CL-CLIP, a multi-agent system that extends beyond the visual capacities of current state-of-the-art MLLMs to capture subtle pharmacological features, provide interpretable analysis, and facilitate hypothesis generation in pharmacological research.

\subsection{Contrastive Learning}
Contrastive learning is a self-supervised paradigm that learns representations by pulling semantically related pairs closer and pushing unrelated pairs apart in a shared embedding space~\citep{hu2024comprehensive}. In biology, contrastive learning has underpinned several applications, such as single-cell multi-omics integration (scRNA-seq and scATAC-seq)~\citep{liu2025sci2cl}, protein function prediction for classify enzyme activities~\citep{yang2024improved}, drug-target interaction prediction through protein-compound embedding~\citep{singh2023contrastive}. CLIP exemplifies the dual-encoder contrastive paradigm for multi-modal learning, it trains an image encoder and a text encoder so that matched image–text pairs have high cosine similarity while mismatched pairs are pushed apart. By scaling to large, CLIP can produce transferable embeddings that generalize across tasks.

\section{Dataset backgrounds}
\label{apx:datasets}
BBBC021 profiles MCF‑7 cells treated with 38 reference drugs covering 12 mechanisms of action, imaged across up to eight half‑log doses and three channels (DNA, $\beta$‑tubulin, actin) ~\citep{caie2010high}. CPJUMP1 includes 301 small molecules (46 controls) perturbed in U2OS and A549 cells, imaged in five channels (DNA; mitochondria; actin/Golgi/plasma membrane; nucleoli and cytoplasmic RNA; endoplasmic reticulum)~\citep{chandrasekaran2024three}.
RxRx3 assays HUVECs with 1,674 bioactive compounds across eight concentrations and six fluorescence channels to capture dose–response phenotypes~\citep{fay2023rxrx3}. 
\section{Detailed RDKit2D Feature Overview}
\label{apx:RDKit2D}

\begin{table}[H]
\caption{Categorized RDKit2D Descriptors Used in This Study (174 descriptors)}
\centering
\renewcommand{\arraystretch}{1.2}

\resizebox{\textwidth}{!}{%
\begin{tabular}{p{5cm} p{10.5cm}}
\toprule
\textbf{Feature Category} & \textbf{Descriptors} \\
\midrule

Topological and Complexity Descriptors &
BalabanJ, BertzCT, Chi0, Chi0n, Chi0v, Chi1, Chi1n, Chi1v, Chi2n, Chi2v, Chi3n, Chi3v, Chi4n, Chi4v, Ipc, Kappa1, Kappa2, Kappa3 \\

Basic Physicochemical Properties &
MolWt, ExactMolWt, HeavyAtomMolWt, MolLogP, MolMR, LabuteASA, TPSA \\

Atom and Bond Counts &
HeavyAtomCount, NumValenceElectrons, NumRotatableBonds, NumHAcceptors, NumHDonors, NHOHCount, NOCount, NumHeteroatoms, FractionCSP3 \\

Ring Structure Descriptors &
RingCount, NumAromaticRings, NumSaturatedRings, NumAliphaticRings, NumAromaticCarbocycles, NumAromaticHeterocycles, NumSaturatedCarbocycles, NumSaturatedHeterocycles, NumAliphaticCarbocycles, NumAliphaticHeterocycles \\

Electrotopological State (EState) Descriptors &
MaxEStateIndex, MinEStateIndex, MaxAbsEStateIndex, MinAbsEStateIndex \\

VSA (Van der Waals Surface Area) Descriptors &
EState\_VSA1–11, PEOE\_VSA1–14, SMR\_VSA1–10, SlogP\_VSA1–12, VSA\_EState1–10 \\

Fingerprint Density Descriptors &
FpDensityMorgan1, FpDensityMorgan2, FpDensityMorgan3 \\

Fragment-Based Functional Group Descriptors &
fr\_Al\_COO, fr\_Al\_OH, fr\_Al\_OH\_noTert, fr\_ArN, fr\_Ar\_COO, fr\_Ar\_N, fr\_Ar\_NH, fr\_Ar\_OH, fr\_COO, fr\_COO2, fr\_C\_O, fr\_C\_O\_noCOO, fr\_HOCCN, fr\_Imine, fr\_NH0, fr\_NH1, fr\_NH2, fr\_Ndealkylation1, fr\_Ndealkylation2, fr\_Nhpyrrole, fr\_SH, fr\_aldehyde, fr\_alkyl\_carbamate, fr\_alkyl\_halide, fr\_allylic\_oxid, fr\_amide, fr\_amidine, fr\_aniline, fr\_aryl\_methyl, fr\_azo, fr\_benzene, fr\_bicyclic, fr\_dihydropyridine, fr\_epoxide, fr\_ester, fr\_ether, fr\_furan, fr\_halogen, fr\_hdrzine, fr\_imidazole, fr\_imide, fr\_ketone, fr\_ketone\_Topliss, fr\_lactone, fr\_methoxy, fr\_morpholine, fr\_nitrile, fr\_nitro, fr\_nitro\_arom, fr\_nitro\_arom\_nonortho, fr\_para\_hydroxylation, fr\_phenol, fr\_phenol\_noOrthoHbond, fr\_phos\_acid, fr\_phos\_ester, fr\_piperdine, fr\_piperzine, fr\_priamide, fr\_pyridine, fr\_sulfide, fr\_sulfonamd, fr\_sulfone, fr\_thiazole, fr\_thiophene, fr\_unbrch\_alkane, fr\_urea \\

Drug-Likeness Score &
qed \\
\bottomrule
\end{tabular}
}  
\end{table}

\section{Log-Dose Indexing for Serial Dilution}
\label{apx:log_dose_Indexing}
To represent compound concentrations on a consistent and model-friendly scale, we transform raw concentrations into log-scaled step values. This transformation is based on the assumption that concentrations follow a serial dilution protocol in logarithmic space.

Let $C_{\max } \in \mathbb{R}_{>0}$ denote the nominal maximum concentration for a compound, and let $C \in \mathbb{R}_{>0}$ be any intermediate concentration point. In a standard protocol with logarithmic dilution spacing, each dose is reduced by a fixed factor per step. This can be expressed as:
\begin{equation}
    C_k=C_{\max } \cdot 10^{-k \cdot \Delta \log }, \quad k=0,1,2, \ldots
\end{equation}
where $\Delta \log >0$ is the logarithmic step size (in base 10. For example, $\Delta \log =0.5$ corresponds to a ~3.16-fold dilution between adjacent doses, since $10^{-0.5} \approx 0.3162$.

To recover the step index $s(C)$ corresponding to any concentration $C$, we invert the above relation:
\begin{equation}
    \begin{aligned}
C & =C_{\max } \cdot 10^{-s(C) \cdot \Delta \log } \\
\Rightarrow \log _{10}(C) & =\log _{10}\left(C_{\max }\right)-s(C) \cdot \Delta \log \\
\Rightarrow s(C) & =\frac{\log _{10}\left(C_{\max }\right)-\log _{10}(C)}{\Delta \log }
\end{aligned}
\end{equation}
Thus, the log-scaled step transformation is defined as:
\begin{equation}
  s(C):=\frac{\log _{10}\left(C_{\max }\right)-\log _{10}(C)}{\Delta \log }, \quad \Delta \log =0.5
\end{equation}
This representation maps concentrations to a normalized step index in log space, which is more suitable for modeling, especially in contexts where concentration-response relationships are approximately log-linear.

\section{Context-Aware Token Projection Modules}
\label{token_replace_modules}
\begin{algorithm}[H]
\caption{CP-CLIP: Context-Aware Token Projection Modules}
\label{alg:CP-CLIP}
\begin{algorithmic}[1]

\Function{EncodeImage}{$x_{\text{img}}$}
    \State $f_{\text{img}} \gets V(x_{\text{img}})$
    \State \Return $\text{normalize}(f_{\text{img}})$
\EndFunction

\Function{EncodeText}{$x_{\text{txt}}, c, t, e$}
    \State $X \gets \text{TokenEmbedding}(x_{\text{txt}})$
    \If{\texttt{<CONC>} in $x_{\text{txt}}$}
        \State $X[\texttt{<CONC>}] \gets \text{conc\_mlp}(c)$ 
        \Comment{$c \in \mathbb{R}^2$, $\text{conc\_mlp}: \mathbb{R}^2 \rightarrow \mathbb{R}^{d_h} \rightarrow \mathbb{R}^{d}$}
    \EndIf
    \If{\texttt{<TIME>} in $x_{\text{txt}}$}
        \State $X[\texttt{<TIME>}] \gets \text{time\_mlp}(t)$ 
        \Comment{$t \in \mathbb{R}^1$, $\text{time\_mlp}: \mathbb{R}^1 \rightarrow \mathbb{R}^{d_h} \rightarrow \mathbb{R}^{d}$}
    \EndIf
\If{\texttt{<CMPD>} in $x_{\text{txt}}$}
    \State $X[\texttt{<CMPD>}] \gets \text{compound\_mlp}(e)$ 
    \Comment{$e \in \mathbb{R}^{d_{\text{cmp}}}$, $\text{compound\_mlp}: \mathbb{R}^{d_{\text{cmp}}} \rightarrow \mathbb{R}^{d_h} \rightarrow \mathbb{R}^d$}
\EndIf
    \State $X \gets X + \text{PosEmb}(X)$
    \State $f_{\text{txt}} \gets T(X)$
    \State \Return $\text{normalize}(f_{\text{txt}})$
\EndFunction

\end{algorithmic}
\end{algorithm}

\section{Training losses}
\label{apx:losses}
We train the alignment with a symmetric CLIP-style contrastive objective. Specifically, we employ the InfoNCE loss, which encourages matched image-text pairs to have high similarity while contrasting them against all other mismatched pairs in the batch:
\begin{equation}
    \mathcal{L}_{\text {InfoNCE}}=\frac{1}{2 N} \sum_{k=1}^N\left[\ell_{\mathrm{CE}}\left(S_{i \rightarrow t}^{(k, ;)}, y_k\right)+\ell_{\mathrm{CE}}\left(S_{t \rightarrow i}^{(k, ;)}, y_k\right)\right]
\end{equation}

Here, $F_i = [f_i^{(1)}, ..., f_i^{(N)}]^\top \in \mathbb{R}^{N \times d}$ and $F_t = [f_t^{(1)}, ..., f_t^{(N)}]^\top \in \mathbb{R}^{N \times d}$ are the batch of normalized image and text embeddings. The similarity matrices are computed as $S_{i \rightarrow t} = s \cdot F_i F_t^\top \in \mathbb{R}^{N \times N}$. The ground-truth labels $y_k \in\{0,1, \ldots, N-1\}$ indicate the correct matching pair for each sample in the batch. 
$\ell_{\text{CE}}(\cdot, \cdot)$ denotes the standard cross-entropy between the similarity scores and the target labels. 

In our experiments, we additionally compare InfoNCE loss with an alternative loss recently proposed in SigLIP, which simplifies the contrastive objective by directly operate joint embeddings in a shared representation space. 

\begin{equation}
    \mathcal{L}_{\text {SigLIP }}=\frac{1}{N} \sum_{k=1}^N \sum_{j=1}^N-\log \sigma\left(y_{k j} \cdot s \cdot\left\langle f_i^{(k)}, f_t^{(j)}\right\rangle\right)
\end{equation}

Here, $s$ is a learnable temperature parameter. To isolate the effect of the loss function from the model architecture, we apply both loss types within our CP-CLIP framework for a fair comparison.

\section{CellProfiler pipeline}
\label{apx:CellProfiler}
For all DNA channels, we extracted per-cell features using the workflow described in Table~\ref{tab:pipeline_DNA}. This pipeline is specifically optimized for nuclear segmentation and feature extraction, using modules that measure grayscale features like shape, texture, and granularity. These features are particularly suitable for DNA stains. For all non-DNA channels (such as Actin, Tubulin, etc.), we applied a consistent pipeline template described in~\ref{tab:pipeline_Actin}. This workflow is tailored to cytoplasmic or filamentous structures, which differ in spatial organization and image characteristics compared to nuclei.

Some feature modules differ between the two workflows, particularly in how certain parameters are configured. For example, texture features were computed at different spatial scales: for DNA, we used smaller scales (e.g., 3, 5, 7) to capture fine-grained nuclear texture, while for non-DNA channels, larger scales (e.g., 5, 10, 15) were used to capture broader cytoskeletal patterns. Similarly, granularity features and shape descriptors such as Zernike moments were customized to reflect the typical size and morphology of structures in each channel. These differences in pipeline configuration ensure that the measurements are biologically meaningful and adapted to the unique characteristics of each fluorescence channel.

\begin{table}[H]
\centering
\caption{CellProfiler pipeline modules and measured features for DNA channel.}
\label{tab:pipeline_DNA}
\renewcommand{\arraystretch}{1.3}
\small

\begin{tabularx}{\textwidth}{@{}>{\raggedright\arraybackslash}p{4.5cm} >{\raggedright\arraybackslash}X >{\raggedright\arraybackslash}p{5.5cm}}
\toprule
\textbf{Module} & \textbf{Key Settings / Notes} & \textbf{Measured Features} \\
\midrule
\textbf{1. Images} & Load images; filter by: \texttt{isimage}, exclude folders with regex & --- \\
\textbf{2. Metadata} & Extract metadata from filename and folder using regex patterns & \texttt{Plate}, \texttt{Well}, \texttt{Site}, \texttt{ChannelNumber}, \texttt{Date} \\
\textbf{3. NamesAndTypes} & Assign names: \texttt{DNA} (grayscale), \texttt{nuclei\_mask} (objects); match rules: \texttt{file contains "DNA"}, \texttt{file contains "nuclei"} & Image names: \texttt{DNA}, \texttt{mask}; Object names: \texttt{nuclei}, \texttt{Nucleus} \\
\textbf{4. Groups} & Grouping disabled & --- \\
\textbf{5. MeasureImageAreaOccupied} & Measure area of \texttt{nuclei} objects & \texttt{AreaOccupied\_nuclei} \\
\textbf{6. MeasureObjectNeighbors} & Measure neighbors of \texttt{nuclei} within 10 pixels & \texttt{Neighbors\_10px\_Count}, \texttt{Neighbors\_10px\_PercentTouching} \\
\textbf{7. MeasureObjectNeighbors} & Measure neighbors of \texttt{nuclei} within 50 pixels & \texttt{Neighbors\_50px\_Count}, \texttt{Neighbors\_50px\_PercentTouching} \\
\textbf{8. MeasureObjectSizeShape} & Measure \texttt{nuclei}; include Zernike moments and advanced features & Shape: \texttt{Area}, \texttt{Perimeter}, \texttt{Solidity}, \texttt{FormFactor}, etc.; Zernike: \texttt{Zernike\_0\_0} to \texttt{Zernike\_9\_9} \\
\textbf{9. MeasureTexture} & Texture of \texttt{DNA} in \texttt{nuclei}; scales: 3, 5, 7; levels: 256; mode: both image and object & Texture features per scale: \texttt{Contrast}, \texttt{Entropy}, \texttt{Correlation}, etc. \\
\textbf{10. MeasureGranularity} & Granularity of \texttt{DNA} in \texttt{nuclei}; radius = 8, spectrum range = 4 & \texttt{Granularity\_1--4\_DNA\_in\_nuclei} \\
\textbf{11. ExportToSpreadsheet} & Export all features with metadata; output file: \texttt{DATA.csv} with prefix \texttt{Expt\_} & All per-object and per-image features above, including per-image mean/median/std \\
\bottomrule
\end{tabularx}
\end{table}

\begin{table}[H]
\centering
\caption{CellProfiler pipeline modules and measured features for Actin channel.}
\label{tab:pipeline_Actin}
\renewcommand{\arraystretch}{1.3}
\small
\begin{tabularx}{\textwidth}{@{}>{\raggedright\arraybackslash}p{4.3cm} >{\raggedright\arraybackslash}X >{\raggedright\arraybackslash}X@{}}
\toprule
\textbf{Module} & \textbf{Key Settings / Notes} & \textbf{Measured Features} \\
\midrule
\textbf{1. Images} & Load images; filter by: \texttt{isimage}, exclude folders with regex & --- \\
\textbf{2. Metadata} & Extract metadata from filename and folder using regex patterns & \texttt{Plate}, \texttt{Well}, \texttt{Site}, \texttt{ChannelNumber}, \texttt{Date} \\
\textbf{3. NamesAndTypes} & Assign names: \texttt{Actin} (grayscale), \texttt{cell\_mask} (objects); \newline
Match rules: \texttt{file contains "Actin"}, \texttt{file contains "cell"} & 
Image names: \texttt{DNA}, \texttt{mask} \newline
Object names: \texttt{nuclei}, \texttt{Nucleus} \\
\textbf{4. Groups} & Grouping disabled & --- \\
\textbf{5. MeasureImageAreaOccupied} & Measure area of \texttt{cell} objects & \texttt{AreaOccupied\_Cell} \\
\textbf{6. MeasureObjectNeighbors} & Measure neighbors of \texttt{cell} within 10 pixels & 
\texttt{Neighbors\_10px\_Count},
\shortstack[l]{\texttt{Neighbors\_10px\_}\\\texttt{PercentTouching}}
 \\
\textbf{7. MeasureObjectNeighbors} & Measure neighbors of \texttt{cell} within 50 pixels & \texttt{Neighbors\_50px\_Count}, \shortstack[l]{\texttt{Neighbors\_50px\_}\\\texttt{PercentTouching}}
 \\
\textbf{8. MeasureObjectSizeShape} & Measure \texttt{cell}; include Zernike moments and advanced features & Shape: \texttt{Area}, \texttt{Perimeter}, \texttt{Solidity}, \texttt{FormFactor}, \texttt{MaxFeretDiameter}, \texttt{EquivalentDiameter}, etc. \newline Zernike: \texttt{Zernike\_0\_0} to \texttt{Zernike\_9\_9} \\
\textbf{9. MeasureTexture} & Texture of \texttt{Actin} in \texttt{cell}; scales: 3, 5, 7; levels: 256 & Texture features per scale: \texttt{Contrast}, \texttt{Correlation}, \texttt{Entropy}, \texttt{SumEntropy}, \texttt{DifferenceEntropy}, \texttt{InfoMeas1}, \texttt{InfoMeas2} \\
\textbf{10. MeasureGranularity} & Granularity of \texttt{Actin} in \texttt{cell}; radius = 8, spectrum range = 4 & 
\shortstack[l]{\texttt{Granularity\_1--4\_}\\\texttt{Actin\_in\_cell}} \\
\textbf{11. ExportToSpreadsheet} & Export all features with metadata; output file: \texttt{DATA.csv} & All per-object and per-image features above, including per-image mean/median/std \\
\bottomrule
\end{tabularx}
\end{table}

\section{Similarity Performance on Seen Drug Compounds}
\label{tab:seen-drug}
\begin{table}[H]
\caption{Similarity Performance on Seen Drug Compounds}
\centering
\setlength{\tabcolsep}{4pt}
\renewcommand{\arraystretch}{1.2}
\resizebox{\textwidth}{!}{%
\begin{tabular}{c c c c c c c c c c c}
\toprule
Model  & Flindokalner & Racecadotril & AZM475271 & Misoprostol & Trazodone & Orantinib  & Rufinamide & lumiracoxib & BIRB-796 & Methoxsalen\\
\midrule
CLIP ViT-B/16  & $\meanstd{0.486}{0.049}$ & $\meanstd{0.528}{0.009}$ & $\meanstd{0.496}{0.032}$ & $\meanstd{0.437}{0.051}$ & $\meanstd{0.499}{0.036}$ & $\meanstd{0.427}{0.044}$ & $\meanstd{0.500}{0.030}$ & $\meanstd{0.433}{0.042}$ & $\meanstd{0.422}{0.041}$ & $\meanstd{0.440}{0.036}$\\
SigLIP ViT-B/16  & $\meanstd{0.308}{0.088}$  & $\meanstd{0.323}{0.075}$  & $\meanstd{0.209}{0.080}$  & $\meanstd{0.329}{0.077}$  & $\meanstd{0.214}{0.074}$  & $\meanstd{0.322}{0.083}$  & $\meanstd{0.222}{0.068}$  & $\meanstd{0.211}{0.063}$& $\meanstd{0.2407}{0.086}$ & $\meanstd{0.314}{0.073}$ \\
\makecell[c]{CP-CLIP SigLIP-ViT-B/16\\\textit{(descriptor)}} & $\meanstd{0.538}{0.066}$  & $\meanstd{0.539}{0.057}$  & $\meanstd{0.456}{0.040}$  & $\meanstd{0.531}{0.052}$  & $\meanstd{0.448}{0.039}$  & $\meanstd{0.545}{0.046}$  & $\meanstd{0.452}{0.042}$ & $\meanstd{0.448}{0.040}$ &  $\meanstd{0.479}{0.059}$ & $\meanstd{0.525}{0.051}$\\
\makecell[c]{CP-CLIP ViT-B/16\\\textit{(fingerprint)}}  & $\meanstd{0.592}{0.050}$ & $\meanstd{0.598}{0.036}$ & $\meanstd{0.510}{0.045}$ & $\meanstd{0.599}{0.043}$ & $\meanstd{0.510}{0.042}$ & $\meanstd{0.602}{0.040}$ & $\meanstd{0.510}{0.036}$ & $\boldsymbol{\meanstd{0.499}{0.049}}$ & $\meanstd{0.516}{0.036}$ & $\meanstd{0.581}{0.051}$ \\
\makecell[c]{CP-CLIP ViT-B/16\\\textit{(descriptor)}}  & $\meanstd{0.590}{0.052}$ & $\meanstd{0.594}{0.037}$ & $\meanstd{0.510}{0.047}$ & $\meanstd{0.595}{0.047}$ & $\meanstd{0.504}{0.046}$ & $\meanstd{0.596}{0.042}$ & $\boldsymbol{\meanstd{0.511}{0.044}}$ & $\meanstd{0.497}{0.049}$ &  $\boldsymbol{\meanstd{0.525}{0.031}}$ &  $\meanstd{0.573}{0.057}$ \\
\makecell[c]{CP-CLIP ViT-L/16\\\textit{(descriptor)}}  & $\boldsymbol{\meanstd{0.608}{0.057}}$ & $\boldsymbol{\meanstd{0.620}{0.043}}$ & $\boldsymbol{\meanstd{0.511}{0.060}}$ & $\boldsymbol{\meanstd{0.626}{0.039}}$ & $\boldsymbol{\meanstd{0.503}{0.053}}$ & $\boldsymbol{\meanstd{0.626}{0.043}}$ & $\meanstd{0.509}{0.057}$ & $\meanstd{0.496}{0.060}$ & $\meanstd{0.513}{0.050}$ & $\boldsymbol{\meanstd{0.599}{0.064}}$\\
\bottomrule
\end{tabular}
}
\end{table}

\newcolumntype{Y}{>{\centering\arraybackslash}X}

\begin{table}[H]
\centering
\scriptsize 
\caption{Seen drugs similarity averaged score}
\renewcommand{\arraystretch}{1.2}
\begin{tabularx}{\textwidth}{Y Y Y Y Y Y}
\toprule
CLIP ViT-B/16 &
SigLIP ViT-B/16 &
\makecell[c]{CP-CLIP\\ SigLIP-ViT-B/16 \\ \textit{(descriptor)}} &
\makecell[c]{CP-CLIP\\ ViT-B/16 \\ \textit{(fingerprint)}} &
\makecell[c]{CP-CLIP\\ ViT-B/16 \\ \textit{(descriptor)}} &
\makecell[c]{CP-CLIP\\ ViT-L/16 \\ \textit{(descriptor)}} \\
\midrule
0.467 & 0.269 & 0.496 & 0.552 & 0.549 & 0.561 \\
\bottomrule
\end{tabularx}
\end{table}

\section{VISTA-2D fine-tune}
\label{VISTA-2D}

The original VISTA2D model does not consistently achieve accurate segmentation across all fluorescent channels, especially when applied to diverse cell painting datasets. To address this limitation, we fine-tuned the segmentation model using the Cell Painting dataset. Figures below illustrate representative instance segmentation results across different channels and datasets (BBBC021, RxRx1, and CPJUMP, respectively), demonstrating improved mask quality and channel-specific accuracy. Three standard instance segmentation metrics are used to evaluate the fine-tuned model’s instance mask quality on 500 test data, with improvements shown in Table~\ref{tab:mask_eval} :

$\bullet$ \textbf{Intersection over Union (IoU)}
The IoU evaluates the overlap between a predicted instance $P$ and ground truth instance label $T$:
\begin{equation}
  \operatorname{IoU}(P, T)=\frac{|P \cap T|}{|P \cup T|}
\end{equation}
Where $|P \cap T|$ is number of pixels in the intersection of $P$ and $T$.

$\bullet$ \textbf{Aggregated Jaccard Index (AJI)}: The AJI generalizes IoU to an entire image containing multiple instances. It is the ratio of the total number of overlapping pixels between matched ground truth and prediction pairs, to the total number of pixels in their union plus the pixels in all unmatched predicted instances, and can be formulated as:
\begin{equation}
    \mathrm{AJI}(P, T)=\frac{\sum_{i=1}^n\left|T_i \cap P_{\pi(i)}\right|}{\sum_{i=1}^n\left|T_i \cup P_{\pi(i)}\right|+\sum_{j \in U}\left|P_j\right|}
\end{equation}
Where $\pi(i)$ the index mapping that assigns predicted instances align with ground truth instances. $U$ is the set of unmatched predicted instances.

$\bullet$ \textbf{Panoptic Quality (PQ)}: PQ is a metric that jointly evaluates segmentation quality and recognition quality in instance segmentation. It reflects both how accurately the matched segments overlap (IoU) and how well all instances are detected (accounting for false positives and false negatives). PQ rewards correct segmentations while penalizing missing or spurious predictions. PQ can be formulated as:
\begin{equation}
    \mathrm{PQ}(P, T)=\underbrace{\frac{1}{|\mathcal{M}|} \sum_{(p, t) \in \mathcal{M}} \operatorname{IoU}(p, t)}_{\text {Segmentation Quality (SQ) }} \times \underbrace{\frac{|\mathcal{M}|}{|\mathcal{M}|+\frac{1}{2}\left|\mathcal{P}_{\text {unmatched }}\right|+\frac{1}{2}\left|\mathcal{T}_{\text {unmatched }}\right|}}_{\text {Detection Quality (DQ) }}
\end{equation}
Where $\mathcal{M}$ is the number of ground truth pairs, $\mathcal{P}_{\text {unmatched }}$ is unmatched predicted instances (False Positives), $\mathcal{T}_{\text {unmatched }}$ is unmatched ground truth instances (False Negatives).

\begin{table}[H]
\centering
\caption{Instance Mask Evaluation Metrics}
\label{tab:mask_eval}
\begin{tabular}{llll}
\toprule
\textbf{VISTA-2d} & \textbf{IoU} & \textbf{AJI} & \textbf{PQ}\\
\midrule
before fine tune &  0.272 & 0.290 & 0.151\\
after fine tune &  0.824 & 0.791& 0.682\\
\bottomrule
\end{tabular}
\end{table}

\begin{figure}[H]
\begin{center}
\label{seg1}
\includegraphics[width=0.95 \linewidth]{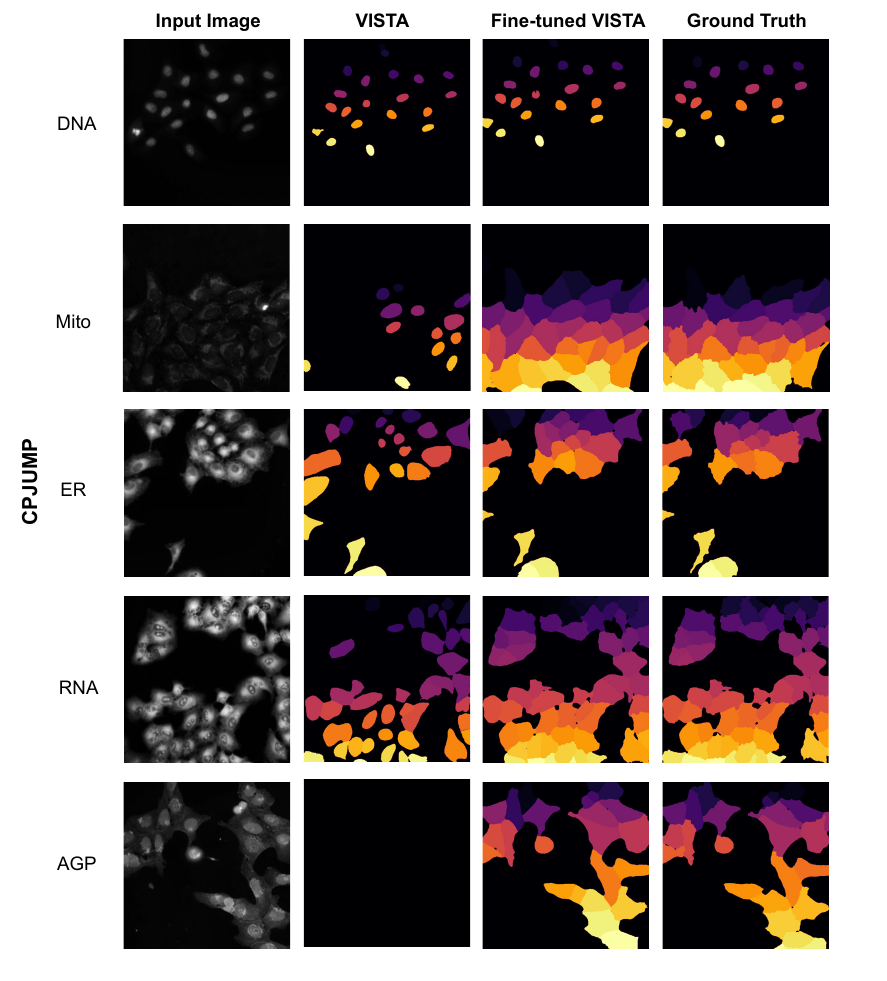}
\end{center}
\caption{Segmentation performance comparison on CP-JUMP dataset across different imaging channels.}
\end{figure}

\begin{figure}[H]
\begin{center}
\includegraphics[width=0.95 \linewidth]{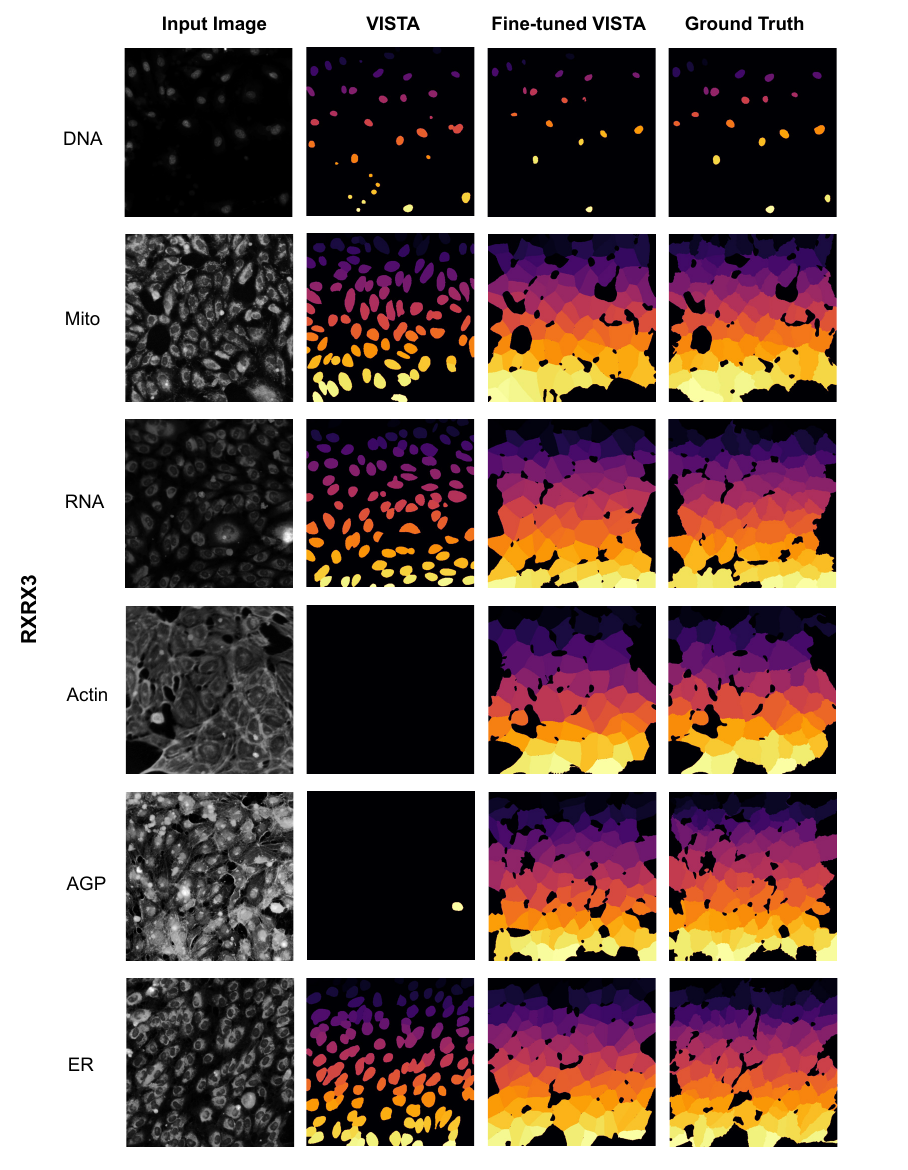}\label{seg2}
\end{center}
\caption{Segmentation performance comparison on RXRX3 dataset across different imaging channels.}
\end{figure}

\begin{figure}[H]
\begin{center}
\includegraphics[width=0.95 \linewidth]{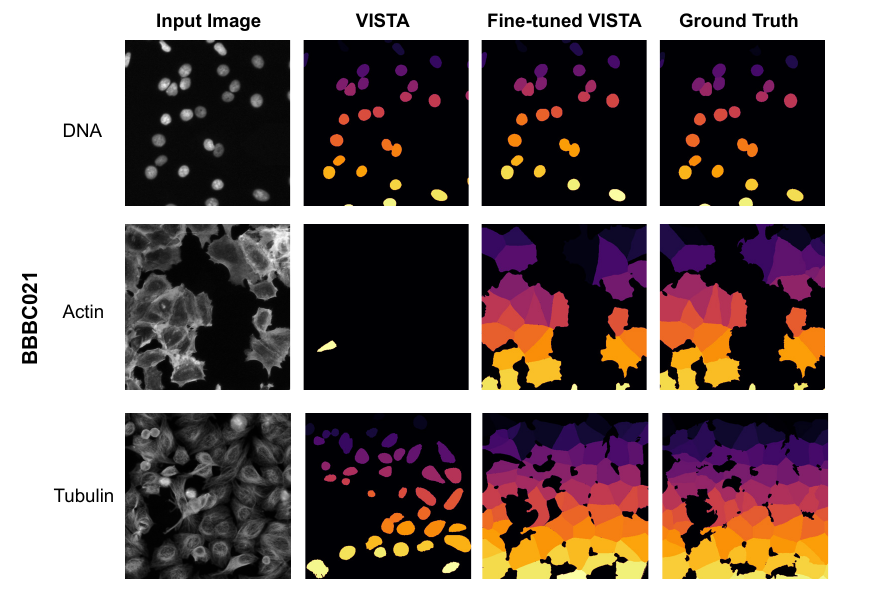}\label{seg3}
\end{center}
\caption{Segmentation performance comparison on BBBC021 dataset across different imaging channels.}
\end{figure}

\section{Dose response examples}
\label{Dose_response}

\begin{figure}[H]
\begin{center}
\includegraphics[width=0.95 \linewidth]{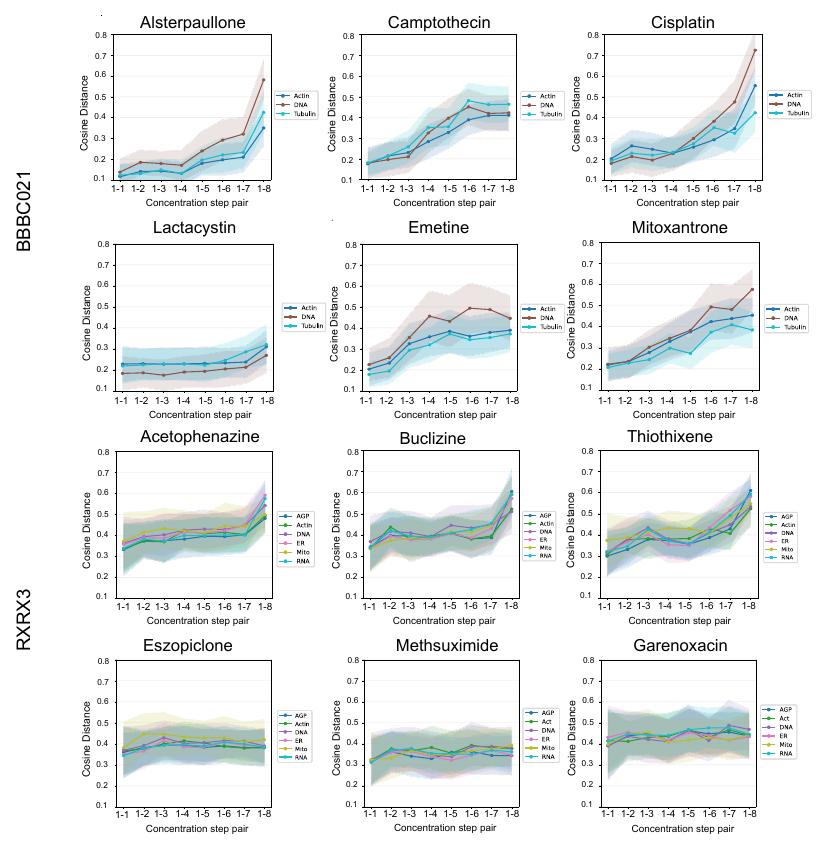}\label{dose_expample}
\end{center}
\caption{Dose–response consistency across compounds in BBBC021 and RxRx3 datasets, measured by cosine distance between CP-CLIP embeddings at different concentration step pairs.}
\end{figure}

To further illustrate the diversity of dose–response behaviors captured by CP-CLIP embeddings, Figure~\ref{dose_expample} shows additional examples from two datasets: BBBC021 and RxRx3 since only the two datasets designed dose scheme based experiments. For each compound, we compute the cosine distance between image embeddings at different concentration levels, focusing on perturbation effects within individual imaging channels.

The x-axis denotes concentration step pairs relative to the first experimental dose. Because different datasets use either fixed or variable half-log concentration series, we normalize the comparisons by indexing each dose level (e.g., 1 for the lowest concentration, 8 for the highest). A label such as "1–2" indicates the cosine distance between embeddings at concentration step 1 and step 2. For example, if the lowest concentration is 0.0001 µM and a half-log step is used, then: step 1 is 0.0001 µM, step 2 is 0.000316 µM, step8 is 0 µM. The cosine distance is computed between embeddings $z_i$ and $z_j$ at two different doses $i$ and $j$, where
\begin{equation}
    d_{i j}=1-\frac{\mathbf{z}_i \cdot \mathbf{z}_j}{\left\|\mathbf{z}_i\right\|\left\|\mathbf{z}_j\right\|}
\end{equation}
The y-axis reflects this cosine distance, providing a quantitative measure of morphological difference between two concentrations. A rising trend along the x-axis indicates increasing morphological divergence from the baseline as concentration increases, which indicating a hallmark of a dose-dependent phenotype. Sharp trajectories are observed for drugs such as Alsterpaullone, Camptothecin, Cisplatin, Emetine, Mitoxantrone, Acetophenazine, Buclizine, and Thiothixene, which are also consistent with their known mechanisms. In contrast, compounds such as Eszopiclone and Methsuximide produce more stable embeddings across doses, suggesting limited morphological response. These visualizations provide additional support for the claim that CP-CLIP embeddings can sensitively capture dose-dependent morphological variation across diverse chemical perturbations.

\section{Statistical Evidence Synthesizer Equations}
\label{apx:statistics_synthesizer}
\begin{table}[H]
\centering
\caption{Summary of statistical parameters for image}
\label{tab:statistics}
\begin{tabular}{lll}
\toprule
\textbf{Parameter Name} & \textbf{Expression} & \textbf{Variable Description} \\
\midrule
\texttt{n\_control} & $|a|$ & $a$: Number of cells from the control group \\
\texttt{n\_perturb} & $|b|$ & $b$: Number of cells from the perturbation group \\
\bottomrule
\end{tabular}
\end{table}

\begin{table}[H]
\centering
\caption{Summary of statistical parameters for each feature metric and their definitions}
\label{tab:statistics}
\resizebox{\textwidth}{!}{%
\begin{tabular}{lll}
\toprule
\textbf{Parameter Name} & \textbf{Expression} & \textbf{Variable Description} \\
\midrule
\texttt{median\_control} & $\mathrm{median}(a)$ & $\mathrm{median}$: Median of $a$ \\
\texttt{median\_perturb} & $\mathrm{median}(b)$ & $\mathrm{median}$: Median of $b$ \\
\texttt{mad\_control} & $\mathrm{median}(|a - \mathrm{median}(a)|)$ & MAD: Median absolute deviation of $a$ \\
\texttt{mad\_perturb} & $\mathrm{median}(|b - \mathrm{median}(b)|)$ & MAD: Median absolute deviation of $b$ \\
\texttt{p10\_control} & $Q_a(0.10)$ & $Q_a(p)$: $p$-th quantile of control group $a$ \\
\texttt{p25\_control} & $Q_a(0.25)$ & Same as above \\
\texttt{p50\_control} & $Q_a(0.50)$ & Same as above \\
\texttt{p75\_control} & $Q_a(0.75)$ & Same as above \\
\texttt{p90\_control} & $Q_a(0.90)$ & Same as above \\
\texttt{p10\_perturb} & $Q_b(0.10)$ & $Q_b(p)$: $p$-th quantile of perturbation group $b$ \\
\texttt{p25\_perturb} & $Q_b(0.25)$ & Same as above \\
\texttt{p50\_perturb} & $Q_b(0.50)$ & Same as above \\
\texttt{p75\_perturb} & $Q_b(0.75)$ & Same as above \\
\texttt{p90\_perturb} & $Q_b(0.90)$ & Same as above \\
\texttt{delta\_median} & $\mathrm{median}(b) - \mathrm{median}(a)$ & Difference in medians between groups \\
\texttt{bootstrap\_ci\_lower} & $\mathrm{CI}_{\mathrm{low}}$ & Lower bound of bootstrap confidence interval \\
\texttt{bootstrap\_ci\_upper} & $\mathrm{CI}_{\mathrm{up}}$ & Upper bound of bootstrap confidence interval \\
\texttt{cliffs\_delta} & $d$ & $d$: Cliff’s delta effect size \\
\texttt{p\_value} & $p$ & $p$: Statistical significance from hypothesis test \\
\bottomrule
\end{tabular}%
}
\end{table}

The lower and upper bounds of the bootstrap confidence interval, denoted as $\mathrm{CI}_{\mathrm{low}}$ and $\mathrm{CI}_{\mathrm{up}}$, estimate the confidence interval of the median difference between control and perturbed sample using the bootstrap resampling method. Specifically, 1000 rounds of bootstrap sampling are performed. It can be computed as:
\begin{equation}
    \mathrm{CI}_{\text {low }}=\text { Percentile }_{2.5}\left(\left\{\delta_i^*\right\}\right)
\end{equation}
\begin{equation}
    \mathrm{CI}_{\mathrm{up}}=\text { Percentile }_{97.5}\left(\left\{\delta_i^*\right\}\right)
\end{equation}

Let $\delta_i^*$ denote the median difference obtained in the $i$-th round of bootstrap resampling, the collection $\left\{\delta_i^*\right\}$ represents the set of median differences obtained from $N$ rounds of bootstrap resampling.

Cliff’s delta is a nonparametric effect size that quantifies the magnitude of difference between two distributions. It is computed as:
\begin{equation}
    d=\frac{1}{|a| |b|} \sum_{i=1}^{n_x} \sum_{j=1}^{n_y}\left[\mathbb{I}\left(x_i>y_j\right)-\mathbb{I}\left(x_i<y_j\right)\right]
\end{equation}

Where $x_i$ denotes the $i$-th sample from the control group, and $y_j$ denotes the $j$-th sample perturbation (or treatment) group. The indicator function $\mathbb{I}(\cdot)$ returns 1 if the condition inside the brackets is true, and 0 otherwise.  Cliff’s delta, which quantifies the degree of difference between the two groups. Its value ranges from $-1$ to $1$, where $d=0$ indicates no difference, $d=1$ indicates the control group has a much bigger value.

The $p$-value corresponds to the result of a two-sided Mann–Whitney U test. It helps assess whether the observed difference could be explained by random variation, under the assumption that the null hypothesis is true. The $p$-value is computed as:
\begin{equation}
    p=2 \cdot\left(1-\Phi\left(\left|\frac{U-\frac{n_a n_b}{2}}{\sqrt{\frac{n_a n_b\left(n_a+n_b+1\right)}{12}}}\right|\right)\right)
\end{equation}
Where $U$ is the Mann–Whitney U statistic, and $n_a$, $n_b$ are the sample sizes of the two groups being compared. The term $\Phi(\cdot)$ denotes the cumulative distribution function (CDF) of the standard normal distribution. The numerator measures the deviation of the observed U value from its expected value under the null hypothesis. This standardization transforms the $U$ statistic into a z-score, which is then used to compute the two-tailed p-value. A small p-value indicates that the observed difference in distributions is unlikely to have occurred by chance.

\section{MLLMs Baseline Details}
\label{apx:MLLM}
\subsection{Methods}
To evaluate the reasoning capability of current mainstream MLLMs on the Cell Painting dataset, we test four API-accessible models: Grok-4, GPT-5, Claude-4-Sonnet, and Gemini-2.5-Pro. The experimental workflow consists of two stages. First, each MLLM performs background knowledge curation as a single preliminary task. The curated information is then used as context for zero-shot VQA across three tasks: the cell line task, the channel task, and the perturbation compound task. During background knowledge curation, the decoding parameters are set to temperature = 0.7 and top-p = 0.95, whereas for VQA they are set to temperature = 1 and top-p = 1 to ensure response stability. All MLLMs are prompted with the same structured instructions specifying the evaluation criteria. In the VQA stage, the models receive both control and perturbation images together with masked textual descriptions. Their task is to select the correct answer from multiple-choice options that include the ground-truth label and to provide both a confidence estimate and a concise rationale. An example prompt is shown below.

In addition to the zero-shot setting described above, we further evaluate a few-shot variant of the same protocol to make the comparison with CP-Agent more conservative. For each of the three tasks (cell line, channel, and perturbation compound), we construct a small visual memory bank consisting of two labeled exemplar image pairs per class (control + perturbation). These exemplars are selected from the training split and are fixed across all MLLMs to ensure comparability. In the few-shot condition, the VQA prompt is augmented with these exemplars: before answering a query, the model is shown the memory bank with the corresponding class labels and is instructed to use these examples as visual references when reasoning about the new control–perturbation pair.

The overall prompting structure, background knowledge curation stage, and decoding parameters remain identical to the zero-shot setup. The only difference is the inclusion of the exemplar memory bank in the VQA stage. As reported in Table~\ref{tab:mllm_fewshot}, few-shot prompting yields modest improvements on the cell line and channel tasks, indicating that current MLLMs can benefit from limited visual grounding. However, performance on the perturbation compound task remains very low, and the models do not exhibit reliable compound-level discrimination despite changes in the prediction distribution. This suggests that the subtle morphological signatures induced by chemical perturbations are difficult for general-purpose MLLMs to acquire from a small number of Cell Painting exemplars.

\subsection{Prompts}
\label{apx:MLLM_prompts}

\begin{figure}[H]
\begin{center}
\includegraphics[width=0.95 \linewidth]{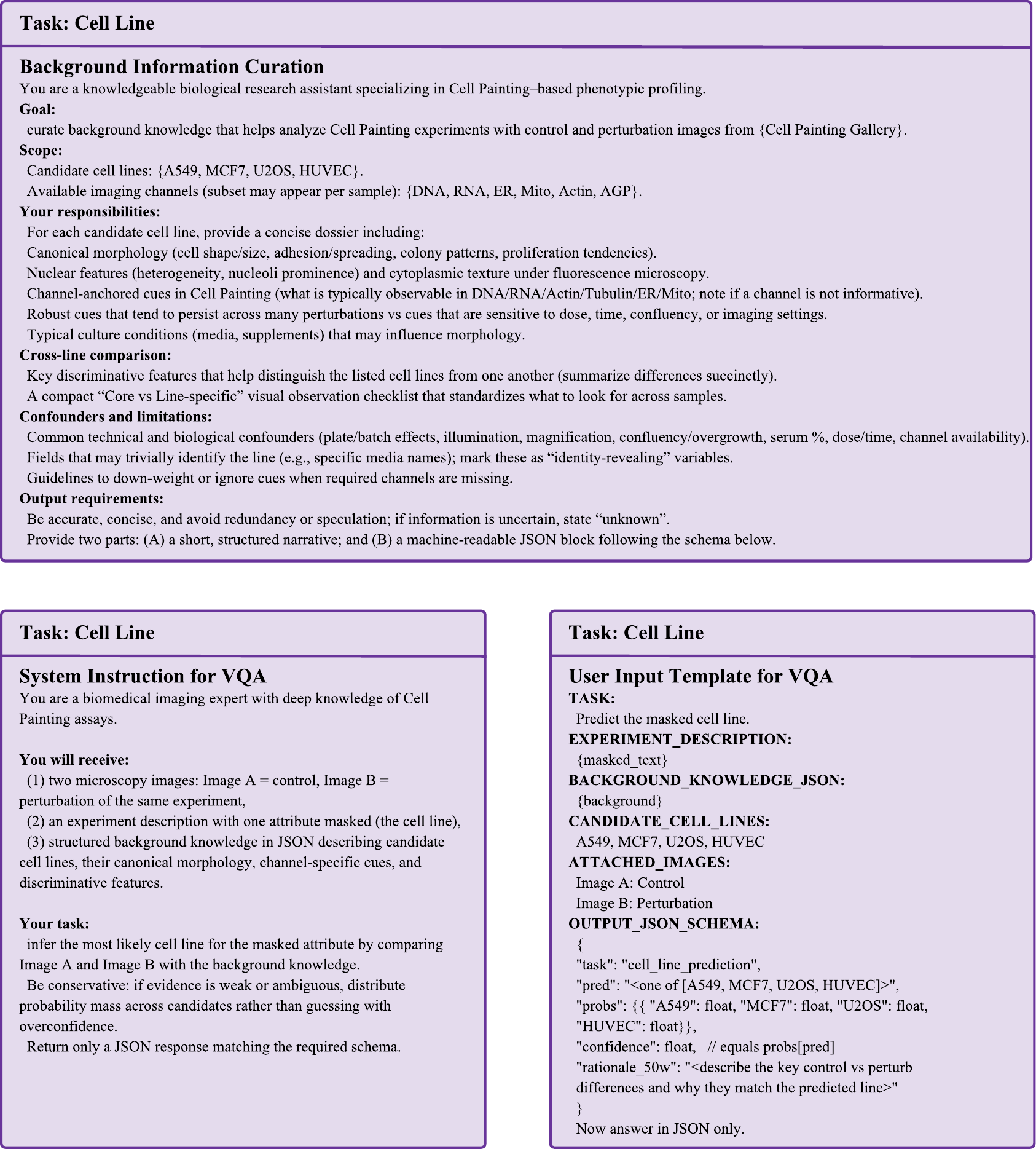}
\end{center}
\end{figure}

\begin{figure}[H]
\begin{center}
\includegraphics[width=0.95 \linewidth]{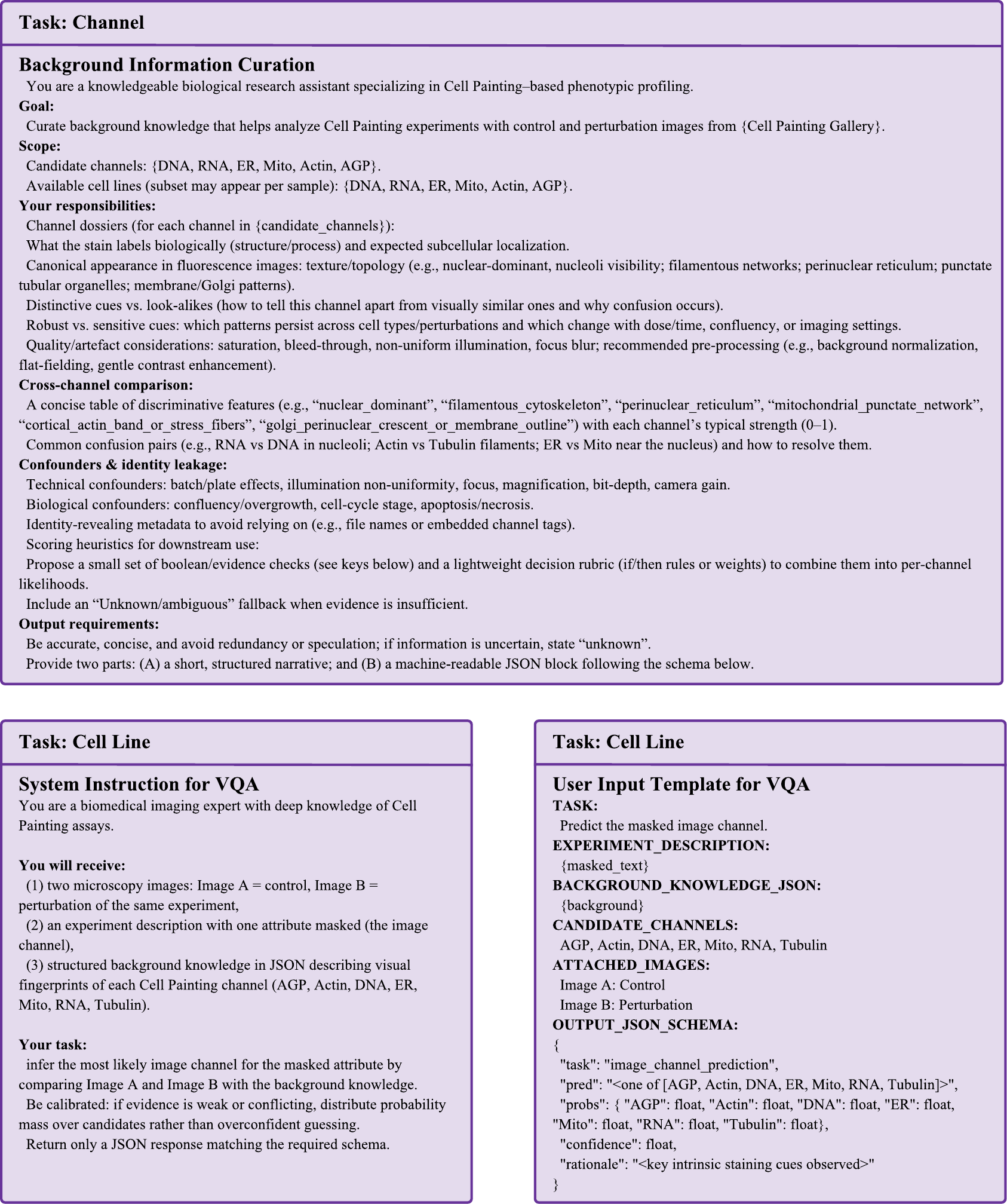}
\end{center}
\end{figure}

\begin{figure}[H]
\begin{center}
\includegraphics[width=0.95 \linewidth]{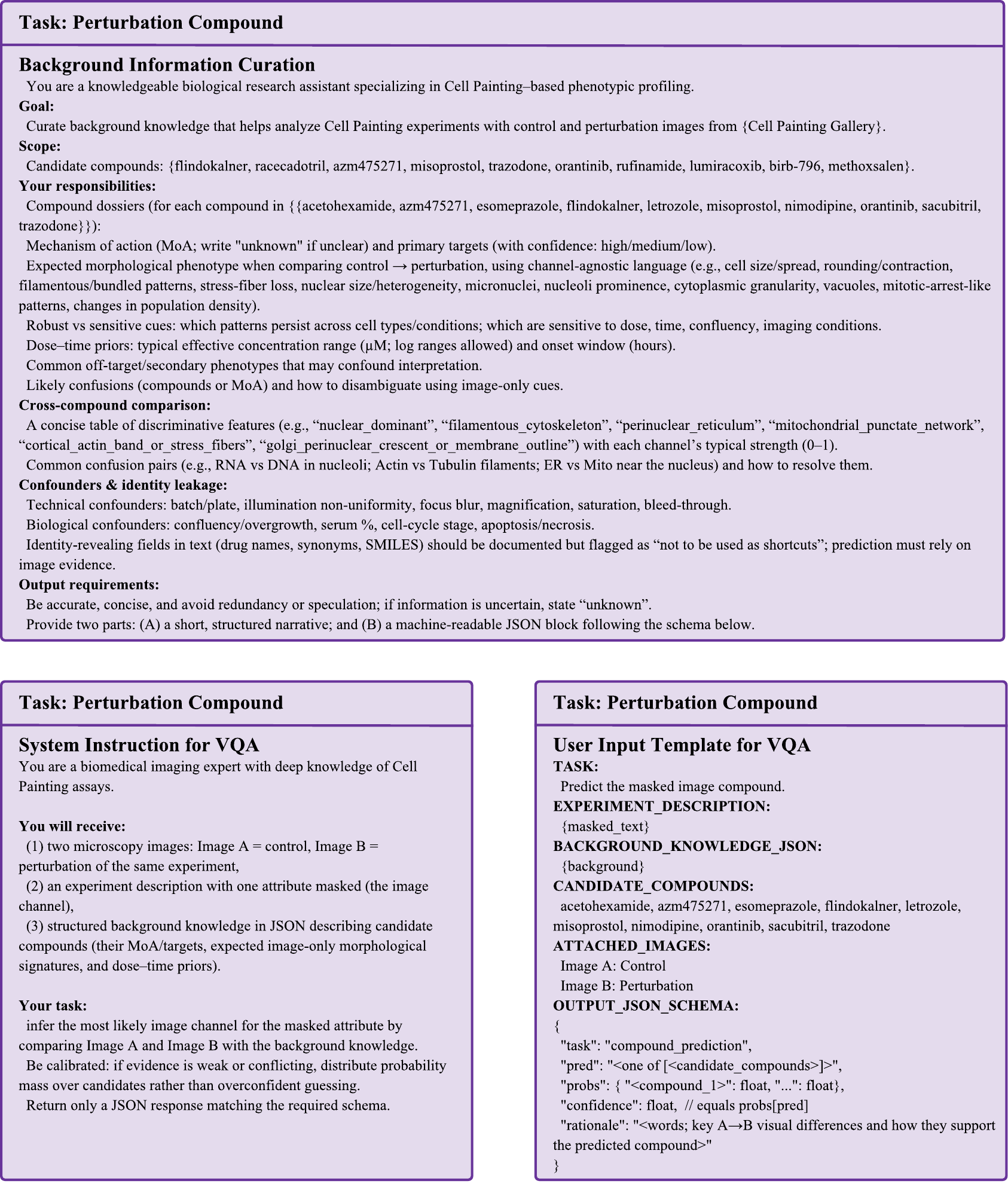}
\end{center}
\end{figure}

\subsection{Detailed Results}
\label{apx: MLLM_performance}

\begin{figure}[H]
\begin{center}
\includegraphics[width=1 \linewidth]{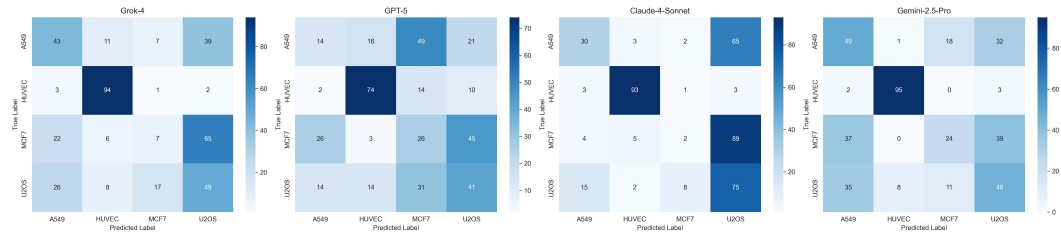}
\end{center}
\vspace{-1.5em}
\caption{Confusion matrix on cell line task.}
\vspace{-1.5em}
\end{figure}

\begin{figure}[H]
\begin{center}
\includegraphics[width=1 \linewidth]{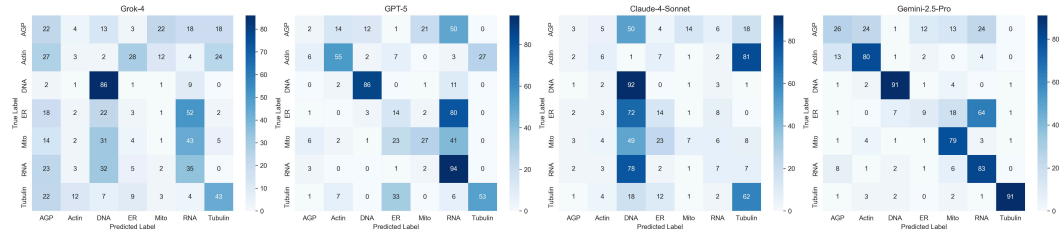}
\end{center}
\vspace{-1.5em}
\caption{Confusion matrix on image channel task.}
\vspace{-1.5em}
\end{figure}

\begin{figure}[H]
\begin{center}
\includegraphics[width=1 \linewidth]{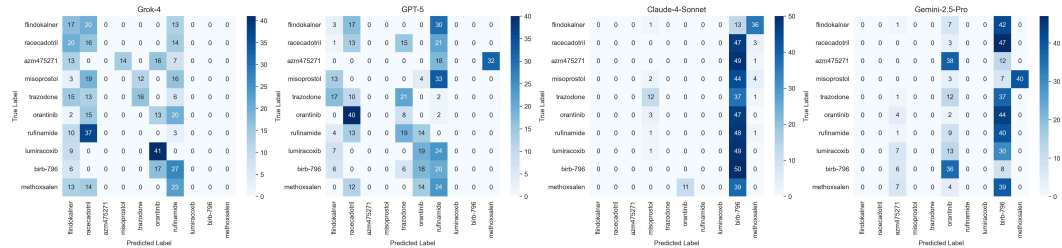}
\end{center}
\vspace{-1.5em}
\caption{Confusion matrix on perturbation compound task.}
\end{figure}

\subsection{Few-shot Results}
\vspace{-2em}
\begin{table}[htbp]
\caption{Few-shot Model performance on classification tasks}
\label{tab:mllm_fewshot}
\centering
\setlength{\tabcolsep}{4pt} 
\renewcommand{\arraystretch}{1.2} 
\resizebox{\textwidth}{!}{%
\begin{tabular}{ccc|*{11}{c}} 
\toprule
\textbf{Model} & \textbf{Cell line} & \textbf{Channel} & \multicolumn{11}{c}{\textbf{Perturbation Compound}} \\
\cmidrule(lr){4-13}
& & & Flindokalner  & Racecadotril & AZM-475271 & Misoprostol & Trazodone & Orantinib & Rufinamide & Lumiracoxib & BIRB-796 & Methoxsalen & Macro-avg \\
\midrule

Grok-4 & 0.515 (+0.067) & 0.260 (+0.032) & 0.224 & 0.184 & 0.0 & 0.0 & 0.390 & 0.176 & 0.034 & 0.000 & 0.0 & 0.000 & 0.101 \\
GPT-5 & 0.440 (+0.063) & 0.510 (+0.071) & 0.0 & 0.0 & 0.066 & 0.0 & 0.115 & 0.000 & 0.079 & 0.000 & 0.088 & 0.000 & 0.035 \\
Claude-4-Sonnet & 0.520 (+0.070) & 0.225 (+0.027) & 0.000 & 0.000 & 0.000 & 0.055 & 0.000 & 0.000 & 0.000 & 0.000 & 0.210 & 0.000 & 0.026 \\
Gemini-2.5-Pro & 0.600 (+0.074) & 0.730 (+0.102) & 0.0 & 0.000 & 0.000 & 0.000 & 0.000 & 0.0 & 0.000 & 0.160 & 0.074 & 0.000 & 0.023 \\

\bottomrule
\end{tabular}
}
\vspace{-1em}
\end{table}

\section{CP-Agent Prompts}
\label{CP_prompt}
\begin{figure}[H]
\begin{center}
\includegraphics[width=1 \linewidth]{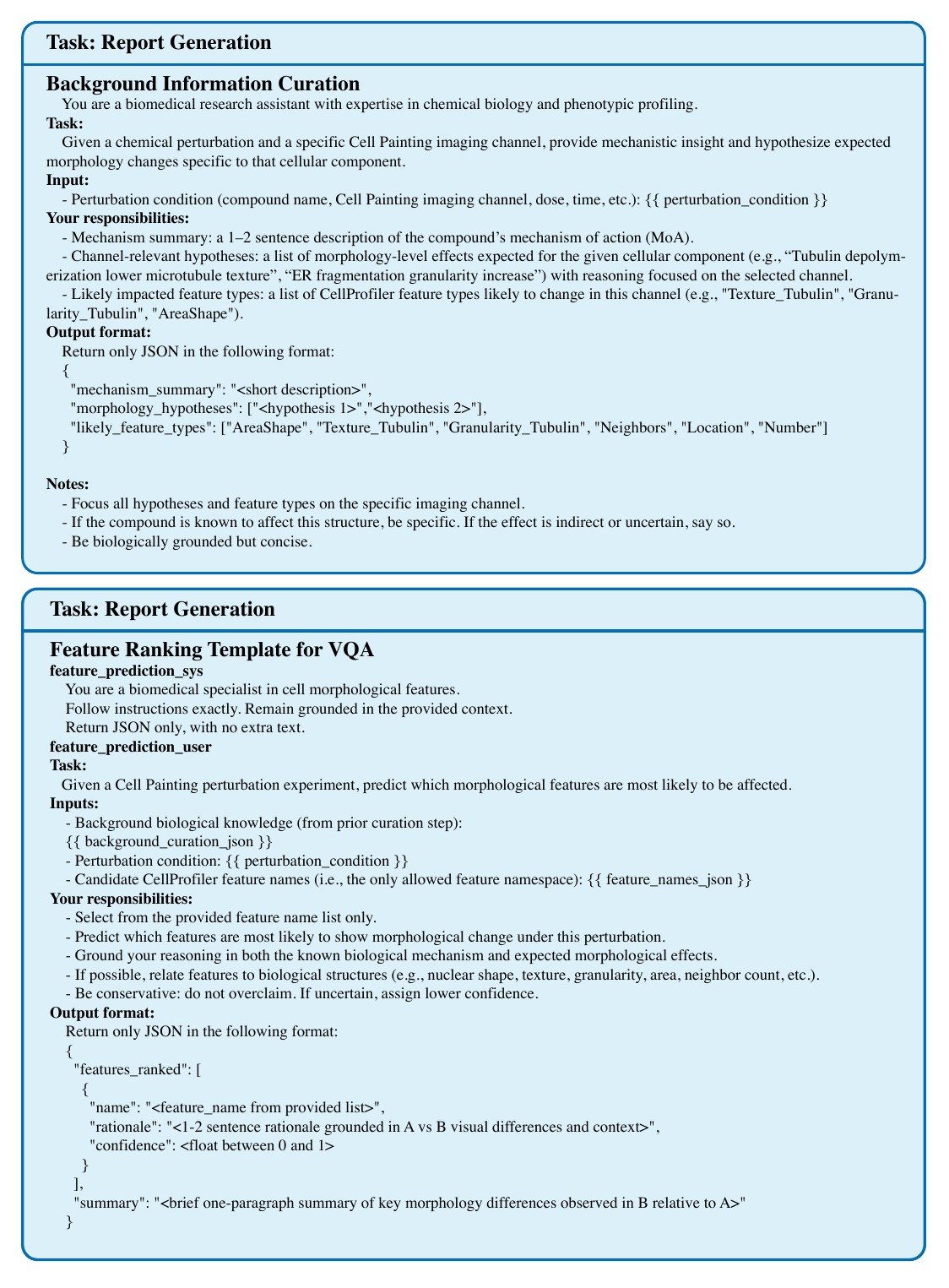}
\end{center}
\vspace{-2em}
\caption{Prompt templates for background curation and feature ranking. }\label{cp-agent_prompt1}
\vspace{-2em}
\end{figure}

\begin{figure}[H]
\begin{center}
\includegraphics[width=1 \linewidth]{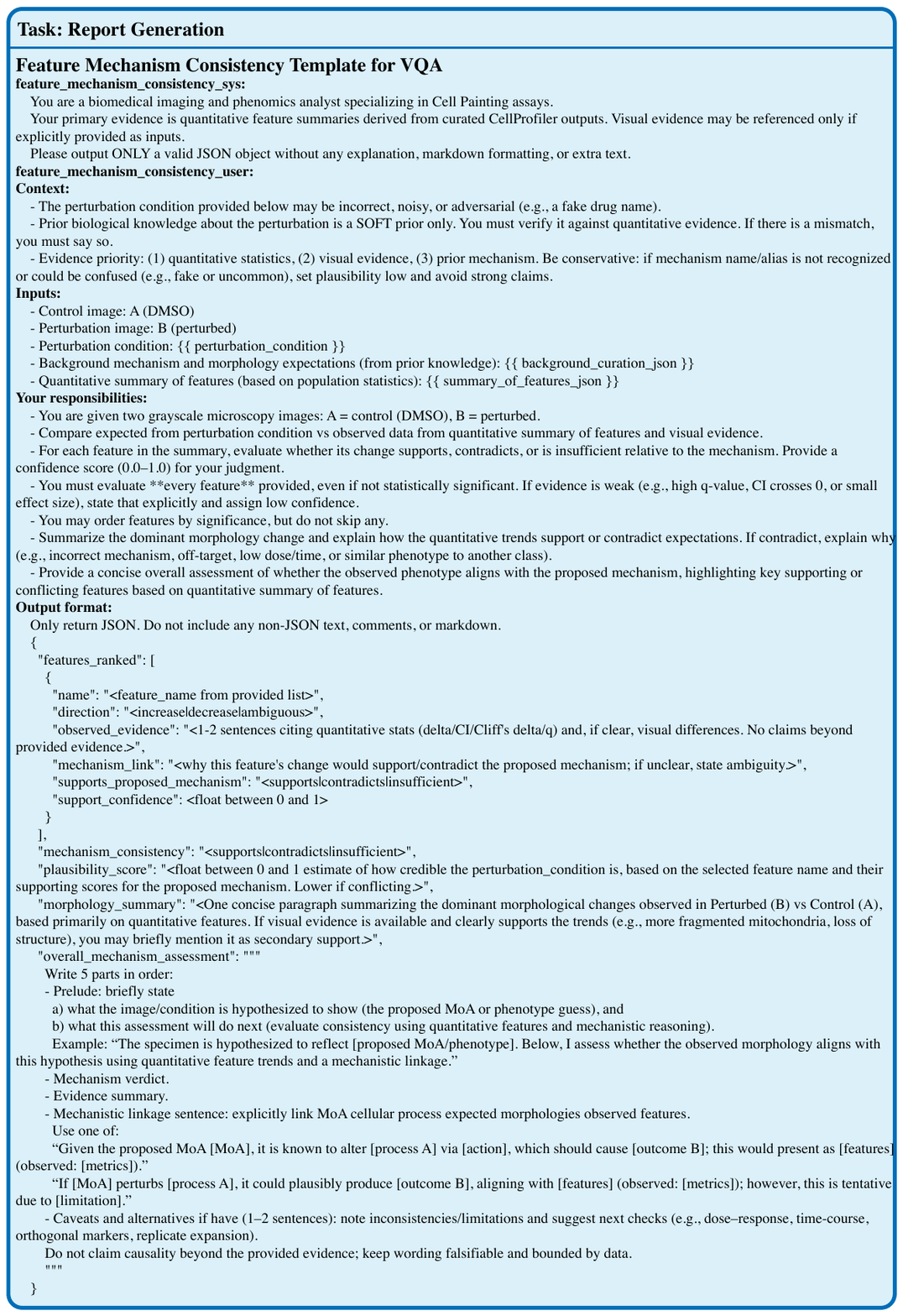}
\end{center}
\vspace{-2em}
\caption{Prompt template for evaluating mechanism-feature consistency in Cell Painting data.}\label{cp-agent_prompt2}
\vspace{-2em}
\end{figure}

The prompts guide the CP-Agent through a multi-step reasoning process to interpret morphological effects of perturbations in Cell Painting data. Figure~\ref{cp-agent_prompt1} introduces two tasks: (1) a background curation step, where the agent synthesizes prior biological knowledge about a compound’s mechanism of action (MoA) and predicts which CellProfiler feature classes are likely to be affected in a specific imaging channel, and (2) a feature ranking task, where individual features are prioritized based on their relevance to the predicted morphological response. Figure~\ref{cp-agent_prompt2} guides the CP-Agent to evaluate whether observed morphological changes under a perturbation are consistent with the proposed mechanism of action (MoA). Using prior biological knowledge and quantitative feature summaries, the agent assesses each feature’s directional change, links it to the expected mechanism, and assigns confidence scores. The agent then provides an overall judgment of mechanism plausibility, highlighting supporting or conflicting evidence. All prompts enforce structured JSON outputs to ensure compatibility with automated downstream analysis and promote reproducibility.

\vspace{-2em}
\label{CP_prompt}

\section{Additional case studies}
\vspace{-1em}
\label{additional_reports}
\subsection{Additional case 1: Taxol in MCF7}
\vspace{-1em}
\begin{figure}[H]
\begin{center}
\includegraphics[width=1 \linewidth]{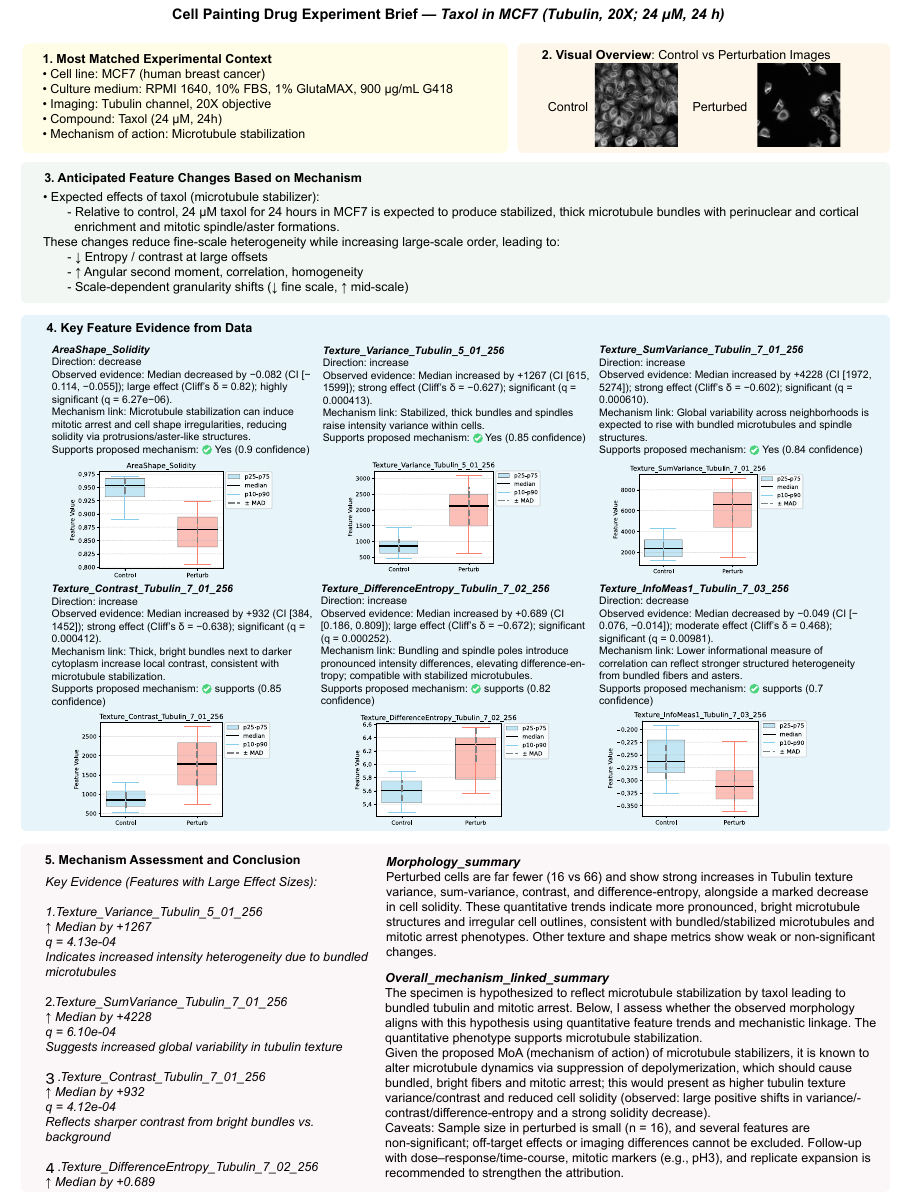}
\end{center}
\end{figure}
\subsection{Additional case 2: Vincristine in MCF7}
\vspace{-1.5em}
\begin{figure}[H]
\begin{center}
\includegraphics[width=1 \linewidth]{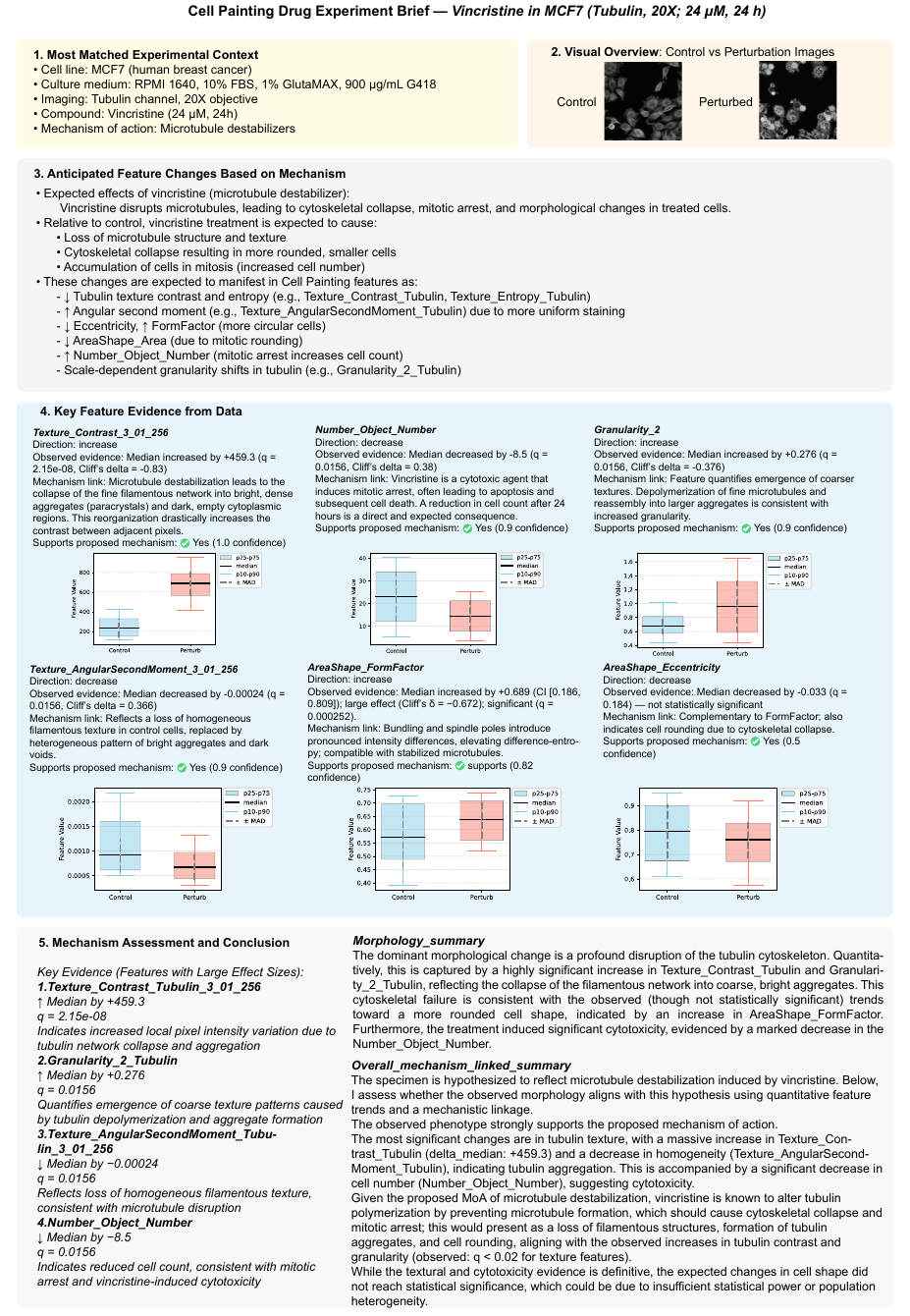}
\end{center}
\end{figure}
\subsection{Additional case 3: Sorbinil in A549}
\vspace{-1.8em}
\begin{figure}[H]
\begin{center}
\includegraphics[width=1 \linewidth]{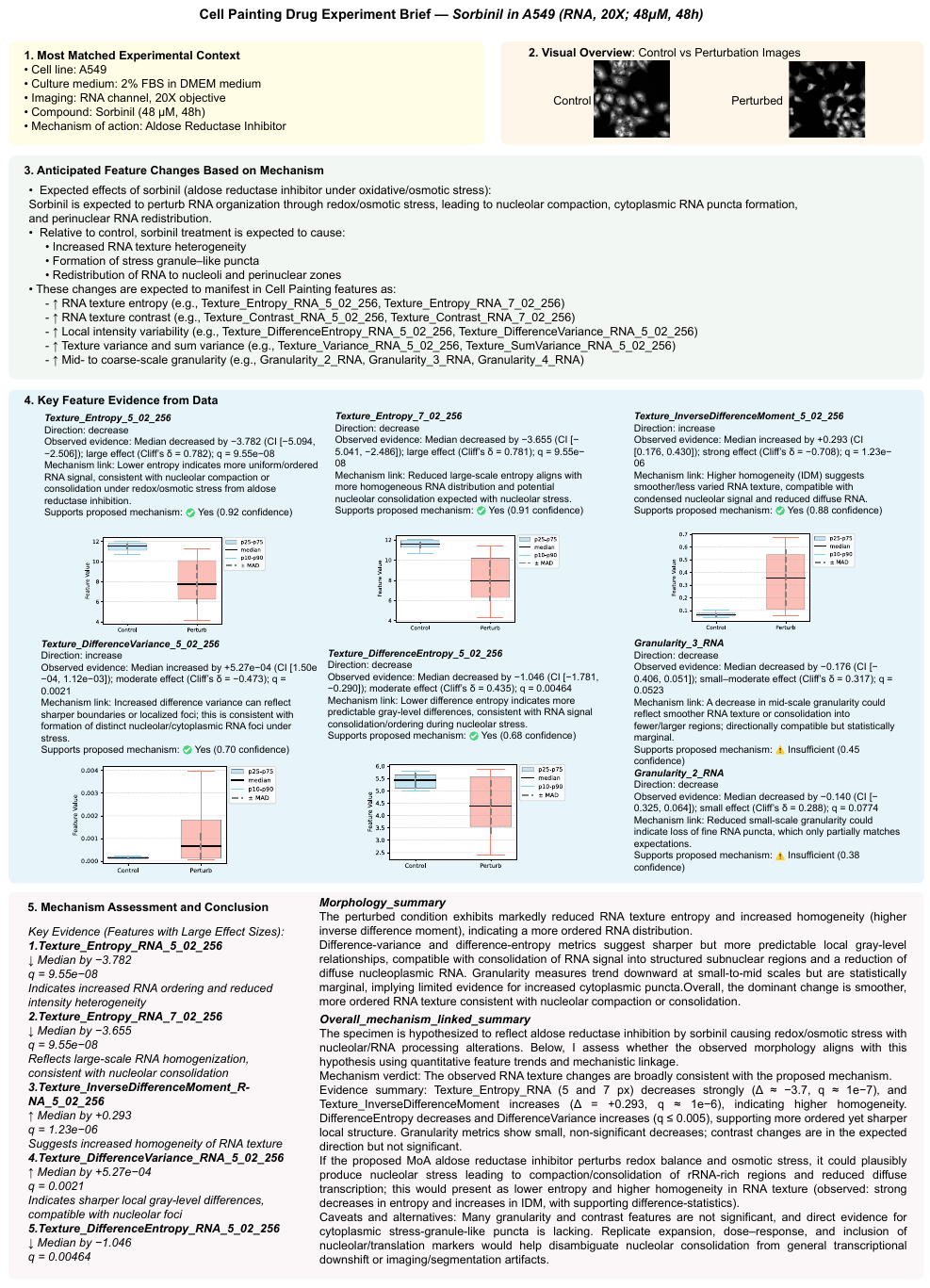}
\end{center}
\end{figure}
\subsection{Additional case 4: BGT226 in HUVEC}
\vspace{-1em}
\begin{figure}[H]
\begin{center}
\includegraphics[width=1 \linewidth]{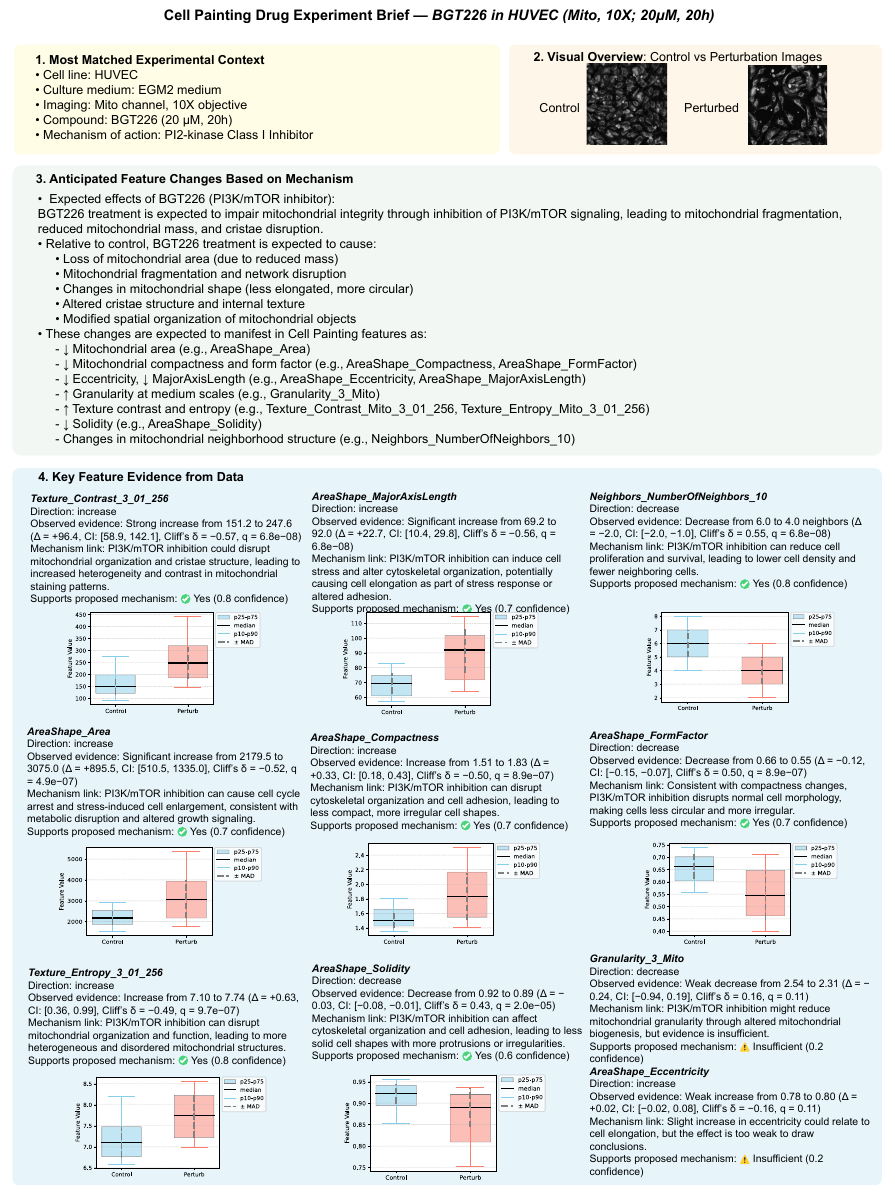}
\end{center}
\end{figure}
\begin{figure}[H]
\begin{center}
\includegraphics[width=1 \linewidth]{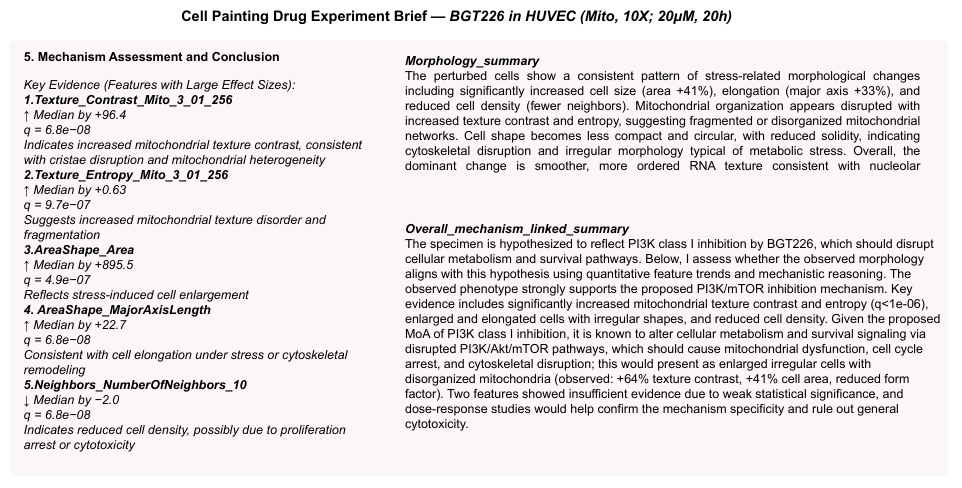}
\end{center}
\end{figure}
\subsection{Additional case 5: AZ841 in MCF7}
\vspace{-1em}
\begin{figure}[H]
\begin{center}
\includegraphics[width=1 \linewidth]{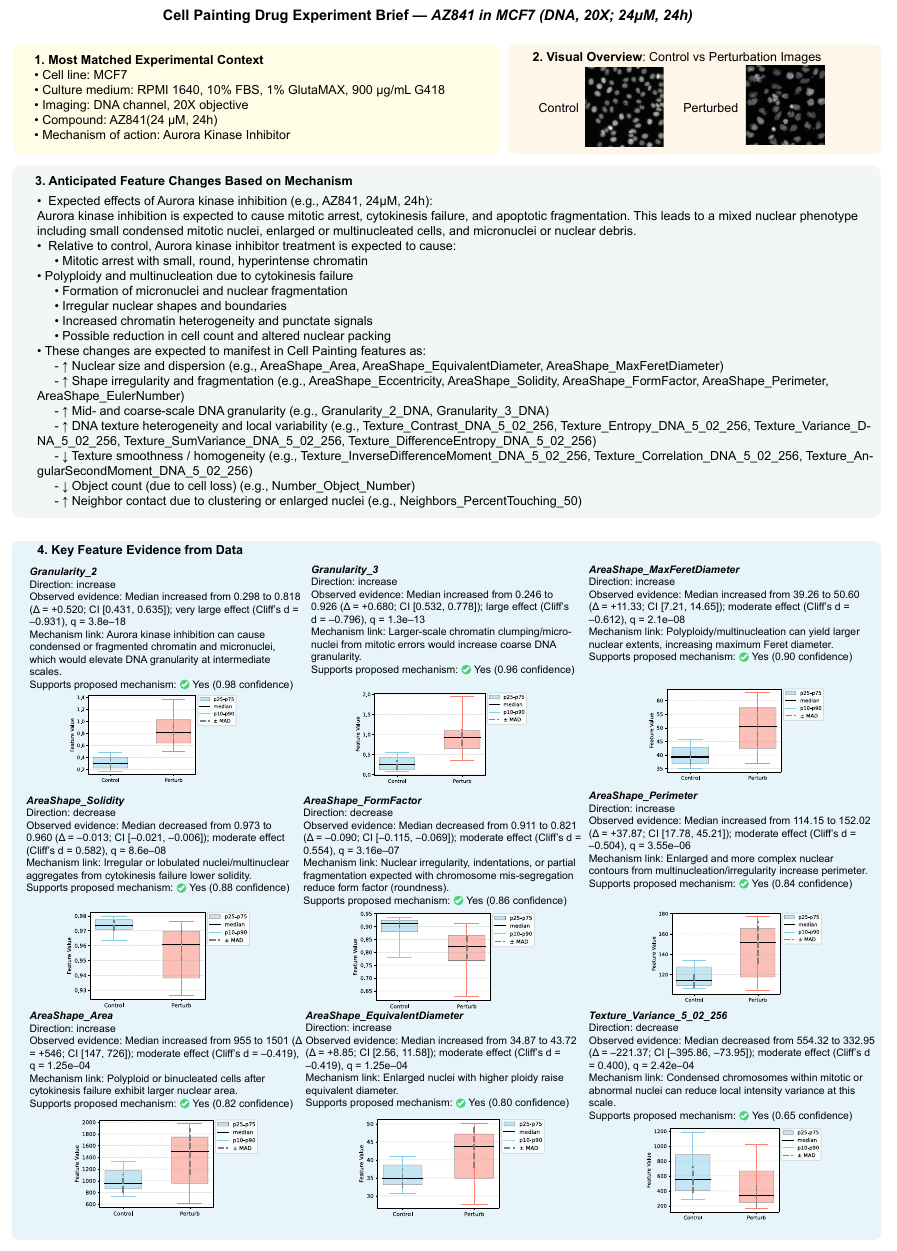}
\end{center}
\end{figure}
\begin{figure}[H]
\begin{center}
\includegraphics[width=1 \linewidth]{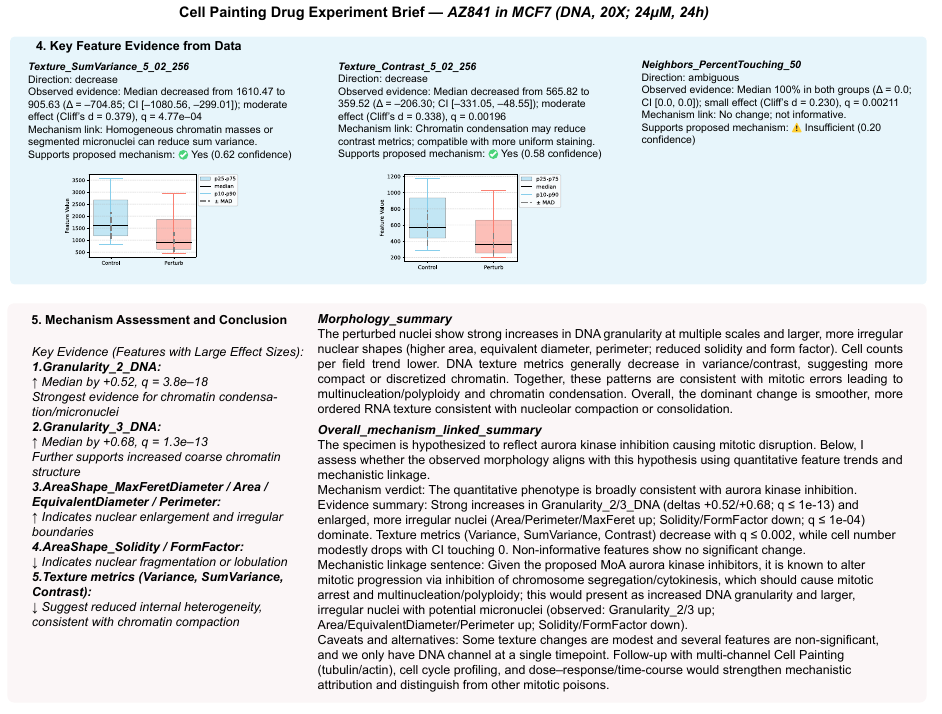}
\end{center}
\end{figure}

\section{Reasoning evaluation criteria}
\label{reasoning_criteria}

\begin{sloppypar}
The survey is designed via Google Form, and can be accessed here: \url{https://docs.google.com/forms/d/e/1FAIpQLSc_W2x6ro6huDANCTaOwc5IGvJ2PUXyvt2zMIKYIlI2npyi3w/viewform?usp=header}
\end{sloppypar}

To facilitate consistent and high-quality responses, we shared the following rubric and example list with participated experts as initial guidance. This framework outlines key criteria for evaluating \textbf{Language Quality } and \textbf{Reasoning Quality} of model-generated explanations in biological tasks. The rubric emphasizes five core aspects of language quality—including accuracy, relevance, coherence, depth, and conciseness, as well as five reasoning quality metrics such as pattern recognition, stepwise reasoning, biological deduction, hypothesis formation, and mechanistic insight. Each criterion is paired with both positive and negative examples to help clarify expectations and common pitfalls.

\subsection{Language Quality Criteria}
\begin{figure}[H]
\begin{center}
\includegraphics[width=1 \linewidth]{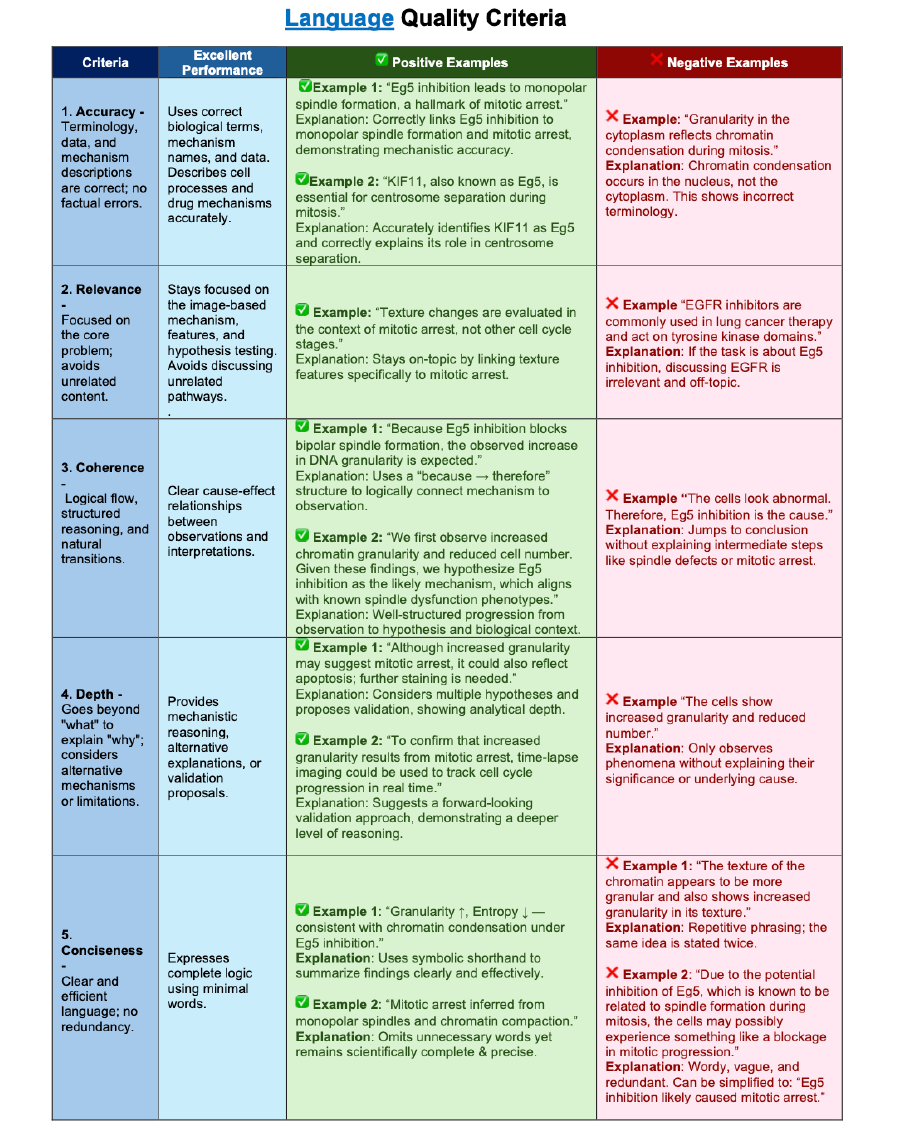}
\end{center}
\caption{Language quality criteria for evaluating CP-Agent generated Cell Painting reports.}
\end{figure}

\subsection{Reasoning Quality Criteria}
\begin{figure}[H]
\begin{center}
\includegraphics[width=1 \linewidth]{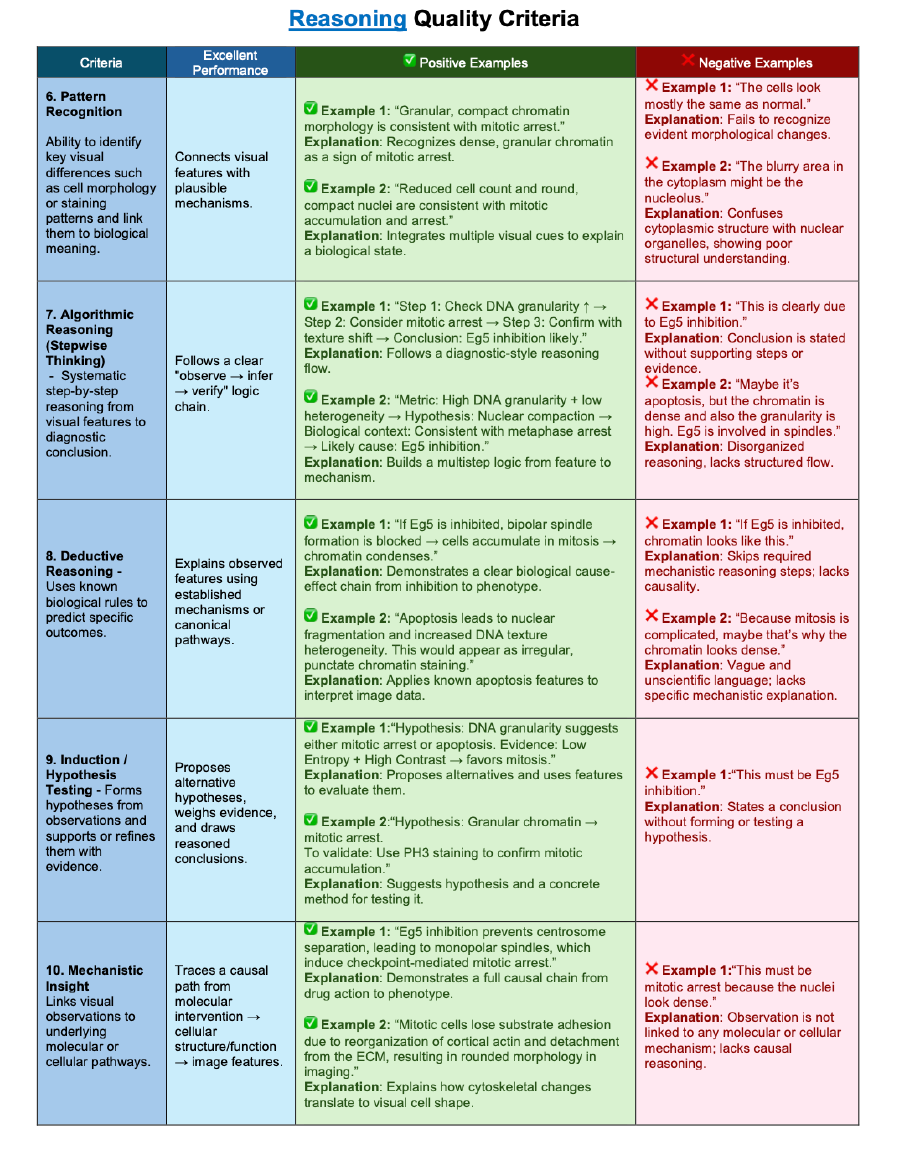}
\end{center}
\caption{Reasoning quality criteria for evaluating CP-Agent generated Cell Painting reports.}
\end{figure}

\section{Expert Ratings of CP-Agent generated Reports Across Language and Reasoning Criteria }

\subsection{Human Expert Assessment of Perturbation Report Quality Across LLMs}
\begin{figure}[H]
\begin{center}
\includegraphics[width=1 \linewidth]{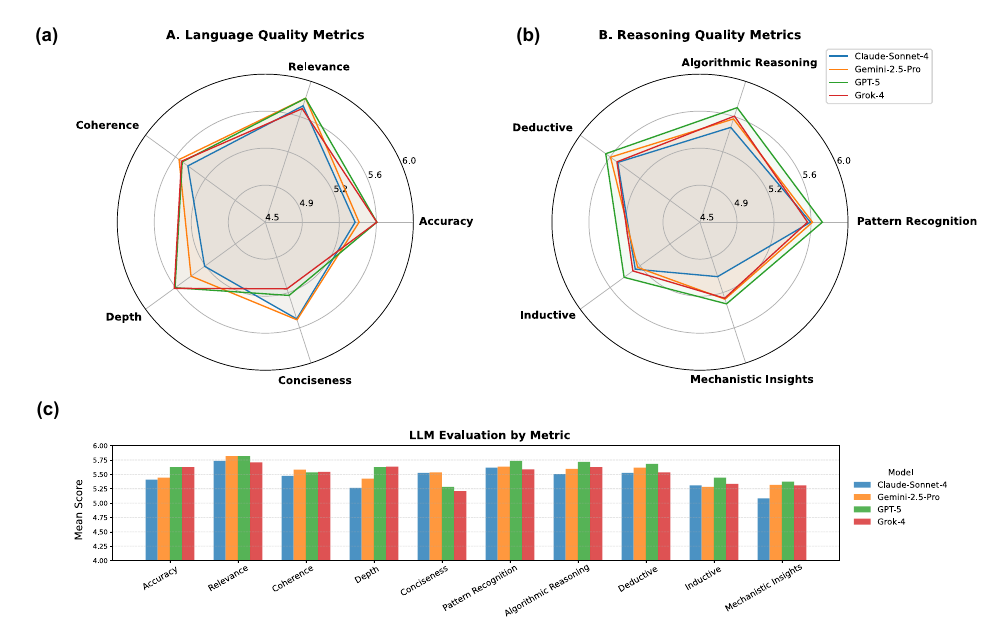}
\end{center}
\caption{\textbf{Expert evaluation of LLM-generated screening reports across language and reasoning dimensions.}(a–b) Mean expert ratings (on a 7-point scale) for language and reasoning quality, based on ten rubric-based evaluation criteria. (c) Bar chart summarizing per-metric mean scores across the four evaluated models: Claude-Sonnet-4 (blue), Gemini-2.5-Pro (orange), GPT-5 (green), and Grok-4 (red). }\label{Appendix_reasoning}
\end{figure}

Figure~\ref{Appendix_reasoning} summarizes expert evaluations across ten rubric criteria, split into five language quality dimensions (Figure~\ref{Appendix_reasoning}a) and five reasoning quality dimensions (Figure~\ref{Appendix_reasoning}b). On average, all four LLMs received high ratings (mostly above 5.0 on a 7-point scale), indicating strong performance in generating biologically grounded screening reports. Among the models, GPT-5 consistently achieved the highest scores across most reasoning metrics—including pattern recognition, algorithmic reasoning, and mechanistic insight—while also maintaining strong language quality. Gemini-2.5-Pro closely followed, particularly excelling in relevance and coherence. Claude-Sonnet-4 underperformed slightly in mechanistic insight and inductive reasoning, indicating slightly weaker performance in higher-order biological inference. Grok-4 showed relatively balanced language quality but lagged slightly in depth and coherence compared to top-performing models. The bar chart (Figure~\ref{Appendix_reasoning}c) further illustrates per-metric mean scores, reinforcing the finding that reasoning dimensions pose a greater challenge than surface-level language quality, especially in tasks requiring mechanistic interpretation and hypothesis generation.

\subsection{Expert Evaluation Consistency Analysis}
To assess the reliability and consistency of expert evaluations used in our study, a comprehensive inter-rater agreement analysis is conducted. The results are detailed in Table~\ref{tab:Inter-rater_Evaluation_full} and summarized in Table~\ref{tab:summary_Inter-rater_Evaluation}. We employed several standard metrics to quantify agreement among expert annotators, including Kendall’s W, standard, deviation, exact agreement percentage, ±1 rating agreement, Mean absolute, difference (MAD). Kendall’s W values ranged from 0.33 to 0.37 across different models, indicating fair agreement among annotators. At the pairwise level, 73.3\% of all rating pairs fell within ±1 point on the 0–7 scale, and the mean absolute difference (MAD) was 1.04, suggesting a high degree of rating consistency. Given the limited number of annotators, bootstrap resampling is employed to compute 95\% confidence intervals for all key reliability metrics, ensuring robust statistical estimation.

Although the expert ratings did not reveal statistically significant differences between models, we observed consistent trends across raters—particularly that GPT-5 achieved marginally higher average scores across most evaluation dimensions. However, these differences were relatively small, implying that all evaluated LLMs were able to generate informative, relevant, and biologically meaningful reports when operating under our CP-Agent generation pipeline.

These findings collectively support the internal consistency and reliability of the expert evaluation process, reinforcing confidence in the qualitative assessments reported in the main text.

\vspace{-1em}
\begin{table}[H]
\centering
\caption{Inter-rater Evaluation for Expert Evaluation}
\resizebox{\textwidth}{!}{%
\label{tab:Inter-rater_Evaluation_full}
\begin{tabular}{llccccccccccccccc}
\toprule
Model & Metric & Kendall\_W & Kendall\_W\_L & Kendall\_W\_U & SD & SD\_L & SD\_U & Exact\% & Exact\_L & Exact\_U & Near\% & Near\_L & Near\_U & MAD & MAD\_L & MAD\_U \\
\midrule
\multirow{10}{*}{Claude-Sonnet-4} & Accuracy & 0.303 & 0.148 & 0.495 & 1.01 & 0.89 & 1.14 & 24.1 & 20.2 & 28.4 & 68.7 & 62.6 & 74.7 & 1.16 & 1.02 & 1.31 \\
& Relevance & 0.349 & 0.194 & 0.548 & 0.95 & 0.85 & 1.04 & 26.0 & 22.4 & 30.4 & 69.4 & 62.1 & 77.1 & 1.11 & 0.96 & 1.24 \\
& Coherence & 0.330 & 0.151 & 0.523 & 0.87 & 0.78 & 0.94 & 30.4 & 25.8 & 36.5 & 75.9 & 69.1 & 82.9 & 0.97 & 0.83 & 1.09 \\
& Depth & 0.291 & 0.122 & 0.484 & 0.91 & 0.85 & 0.97 & 27.2 & 23.5 & 31.3 & 74.5 & 70.8 & 77.6 & 1.03 & 0.95 & 1.13 \\
& Conciseness & 0.244 & 0.106 & 0.434 & 1.05 & 0.90 & 1.23 & 24.8 & 22.0 & 26.9 & 70.4 & 65.0 & 75.6 & 1.17 & 1.03 & 1.34 \\
& Pattern Recognition & 0.312 & 0.138 & 0.520 & 0.89 & 0.77 & 1.04 & 29.9 & 26.2 & 33.2 & 77.1 & 71.2 & 83.3 & 0.99 & 0.86 & 1.14 \\
& Algorithmic Reasoning & 0.368 & 0.153 & 0.650 & 0.88 & 0.82 & 0.94 & 28.1 & 24.9 & 32.0 & 75.9 & 70.9 & 81.1 & 1.00 & 0.90 & 1.09 \\
& Deductive & 0.402 & 0.200 & 0.628 & 0.90 & 0.83 & 0.99 & 29.8 & 26.4 & 33.5 & 74.0 & 68.4 & 78.7 & 1.01 & 0.91 & 1.12 \\
& Inductive & 0.367 & 0.181 & 0.619 & 0.92 & 0.81 & 1.02 & 27.6 & 22.2 & 34.4 & 73.7 & 66.6 & 81.5 & 1.04 & 0.88 & 1.20 \\
& Mechanistic Insights & 0.342 & 0.152 & 0.563 & 1.03 & 0.93 & 1.13 & 23.7 & 21.1 & 26.0 & 66.9 & 60.7 & 72.7 & 1.18 & 1.07 & 1.30 \\
\midrule
\multirow{10}{*}{Gemini-2.5-Pro} & Accuracy & 0.370 & 0.171 & 0.565 & 0.94 & 0.86 & 1.01 & 25.3 & 22.2 & 28.4 & 72.2 & 66.3 & 77.5 & 1.08 & 0.97 & 1.18 \\
& Relevance & 0.409 & 0.191 & 0.664 & 0.94 & 0.87 & 1.00 & 25.0 & 22.4 & 27.7 & 72.3 & 68.2 & 76.5 & 1.08 & 1.00 & 1.16 \\
& Coherence & 0.373 & 0.181 & 0.548 & 0.84 & 0.76 & 0.90 & 29.8 & 24.9 & 35.1 & 79.2 & 74.0 & 84.5 & 0.94 & 0.83 & 1.05 \\
& Depth & 0.300 & 0.119 & 0.518 & 0.85 & 0.80 & 0.90 & 30.4 & 27.1 & 33.8 & 77.0 & 73.4 & 80.9 & 0.95 & 0.88 & 1.03 \\
& Conciseness & 0.302 & 0.151 & 0.472 & 0.89 & 0.83 & 0.94 & 27.0 & 24.7 & 29.5 & 75.2 & 72.0 & 78.2 & 1.02 & 0.95 & 1.08 \\
& Pattern Recognition & 0.359 & 0.205 & 0.526 & 0.84 & 0.74 & 0.91 & 30.5 & 26.1 & 35.6 & 78.3 & 73.0 & 84.8 & 0.94 & 0.82 & 1.05 \\
& Algorithmic Reasoning & 0.381 & 0.179 & 0.597 & 0.81 & 0.73 & 0.90 & 31.9 & 27.9 & 35.6 & 80.5 & 74.4 & 86.8 & 0.90 & 0.79 & 1.02 \\
& Deductive & 0.412 & 0.207 & 0.682 & 0.89 & 0.79 & 1.00 & 31.6 & 26.6 & 36.4 & 75.4 & 68.8 & 81.1 & 0.99 & 0.86 & 1.13 \\
& Inductive & 0.395 & 0.174 & 0.651 & 0.93 & 0.87 & 0.99 & 29.6 & 24.9 & 35.2 & 68.7 & 63.6 & 74.1 & 1.06 & 0.96 & 1.14 \\
& Mechanistic Insights & 0.304 & 0.168 & 0.499 & 0.93 & 0.84 & 1.03 & 28.3 & 23.3 & 33.4 & 71.9 & 65.2 & 78.7 & 1.06 & 0.93 & 1.19 \\
\midrule
\multirow{10}{*}{GPT-5} & Accuracy & 0.337 & 0.175 & 0.504 & 0.95 & 0.86 & 1.04 & 27.2 & 22.4 & 31.8 & 72.1 & 66.5 & 77.3 & 1.08 & 0.96 & 1.21 \\
& Relevance & 0.354 & 0.198 & 0.533 & 0.90 & 0.84 & 0.98 & 27.7 & 24.7 & 31.1 & 70.4 & 65.3 & 75.4 & 1.05 & 0.96 & 1.15 \\
& Coherence & 0.357 & 0.165 & 0.614 & 0.89 & 0.83 & 0.95 & 30.8 & 25.6 & 37.5 & 73.7 & 69.6 & 77.3 & 0.98 & 0.89 & 1.07 \\
& Depth & 0.307 & 0.139 & 0.526 & 0.92 & 0.83 & 1.00 & 29.6 & 25.3 & 34.0 & 71.3 & 63.8 & 78.0 & 1.04 & 0.92 & 1.15 \\
& Conciseness & 0.303 & 0.140 & 0.469 & 0.97 & 0.83 & 1.16 & 28.2 & 24.0 & 32.0 & 71.5 & 63.5 & 77.6 & 1.08 & 0.94 & 1.28 \\
& Pattern Recognition & 0.360 & 0.207 & 0.540 & 0.81 & 0.72 & 0.89 & 32.7 & 28.0 & 38.5 & 79.6 & 74.4 & 84.7 & 0.91 & 0.78 & 1.02 \\
& Algorithmic Reasoning & 0.422 & 0.245 & 0.609 & 0.86 & 0.73 & 0.99 & 30.6 & 24.5 & 37.2 & 78.3 & 69.3 & 86.7 & 0.96 & 0.79 & 1.15 \\
& Deductive & 0.494 & 0.306 & 0.669 & 0.92 & 0.87 & 0.98 & 25.8 & 23.5 & 28.2 & 72.1 & 68.2 & 75.7 & 1.07 & 1.00 & 1.15 \\
& Inductive & 0.424 & 0.254 & 0.609 & 0.95 & 0.91 & 0.99 & 24.5 & 22.2 & 27.1 & 68.7 & 66.5 & 70.7 & 1.12 & 1.07 & 1.17 \\
& Mechanistic Insights & 0.358 & 0.204 & 0.521 & 0.96 & 0.88 & 1.04 & 28.2 & 23.1 & 34.2 & 67.1 & 62.4 & 72.5 & 1.11 & 0.98 & 1.22 \\
\midrule
\multirow{10}{*}{Grok-4} & Accuracy & 0.278 & 0.120 & 0.493 & 0.93 & 0.81 & 1.06 & 27.8 & 23.7 & 32.2 & 73.3 & 66.4 & 80.2 & 1.06 & 0.91 & 1.21 \\
& Relevance & 0.403 & 0.214 & 0.633 & 0.92 & 0.84 & 0.98 & 25.7 & 22.9 & 29.1 & 73.4 & 68.5 & 79.0 & 1.05 & 0.94 & 1.15 \\
& Coherence & 0.367 & 0.199 & 0.586 & 0.81 & 0.68 & 0.94 & 35.6 & 31.2 & 39.7 & 81.1 & 73.8 & 88.7 & 0.87 & 0.73 & 1.03 \\
& Depth & 0.300 & 0.136 & 0.503 & 0.97 & 0.87 & 1.07 & 25.8 & 23.5 & 28.1 & 70.7 & 65.0 & 75.6 & 1.10 & 1.00 & 1.23 \\
& Conciseness & 0.265 & 0.104 & 0.454 & 1.04 & 0.91 & 1.16 & 22.9 & 19.3 & 26.7 & 67.1 & 60.9 & 73.8 & 1.20 & 1.04 & 1.34 \\
& Pattern Recognition & 0.359 & 0.171 & 0.582 & 0.90 & 0.81 & 1.00 & 29.5 & 25.3 & 33.3 & 73.7 & 67.1 & 78.8 & 1.01 & 0.90 & 1.16 \\
& Algorithmic Reasoning & 0.399 & 0.236 & 0.593 & 0.85 & 0.69 & 0.99 & 33.7 & 26.2 & 41.5 & 79.7 & 71.4 & 88.6 & 0.92 & 0.73 & 1.11 \\
& Deductive & 0.498 & 0.279 & 0.700 & 0.89 & 0.81 & 0.98 & 28.8 & 24.4 & 32.5 & 76.3 & 71.6 & 79.8 & 1.00 & 0.91 & 1.11 \\
& Inductive & 0.483 & 0.270 & 0.677 & 0.98 & 0.91 & 1.05 & 25.4 & 22.5 & 28.4 & 69.0 & 64.4 & 73.5 & 1.13 & 1.04 & 1.23 \\
& Mechanistic Insights & 0.323 & 0.149 & 0.519 & 0.99 & 0.86 & 1.11 & 29.1 & 26.0 & 32.4 & 66.6 & 58.9 & 74.7 & 1.12 & 0.97 & 1.27 \\
\bottomrule
\end{tabular}%
}
\end{table}

\begin{table}[H]
\centering
\caption{Summary of Inter-rater Evaluations Across LLMs}
\small
\label{tab:summary_Inter-rater_Evaluation}
\begin{tabular}{lcccc}
\toprule
\textbf{Model} & \textbf{Kendall\_W} & \textbf{Score\_Avg} & \textbf{Score\_Std} & \textbf{Near\_Agreement \%} \\
\midrule
Claude-Sonnet-4 & 0.33 & 5.45 & 0.94 & 72.65 \\
Gemini-2.5-Pro & 0.36 & 5.53 & 0.89 & 75.07 \\
GPT-5 & 0.37 & 5.59 & 0.91 & 72.48 \\
Grok-4 & 0.37 & 5.51 & 0.93 & 73.09 \\
\bottomrule
\end{tabular}
\end{table}

\section{Robustness Evaluation: FeatRank Agent and ReportGen Agent}

The reproducibility of CPAgent’s reasoning outputs is primarily influenced by the temperature parameter in the large language models (LLMs) since the perturbation codition is infered from CP-CLIP, which makes the conclusion of the report deterministic already. In our pipeline, following the initial pretrained CLIP model, there are two LLM modules, the 'FeatRank Agent', which ranks CellProfiler extracted morphology features, and 'ReportGen Agent', which generates natural language reports based on the ranked features and contextual information. To ensure that the feature ranking step is as deterministic as possible, we set the temperature of the FeatRank Agent to 0. To systematically evaluate the reproducibility, we designed five experiments with varying temperature settings, the ReportGen Agent's temperature was set ranging from 0.0 to 1.0, while keeping the FeatRank Agent temperature fixed at 0.0. In each setting, we repeated the pipeline 30 times on the same input samples and analyzed the consistency of the selected features and generated reports.
\subsection*{System Prompt}

You are a scientific evaluator specialized in assessing corpora of mechanism assessment reports derived from Cell Painting assays.

Your task is to evaluate the internal consistency of a set of 30 mechanism assessment reports. Each report analyzes how observed morphological features align with a hypothesized mechanism of action for a specific chemical perturbation.

You are not evaluating the scientific accuracy or biological correctness of any individual report. Your focus is on how mutually consistent the reports are with each other in terms of:
\begin{itemize}
  \item Scientific focus and biological reasoning style
  \item Use of technical terminology and feature names
  \item Presentation structure and rhetorical flow
\end{itemize}

You will be provided with the full set of reports or representative excerpts. Based on this, you will assign scores for each consistency dimension and briefly justify your evaluation.

Return only the JSON structure specified in the user prompt. Do not include any commentary or additional explanation.

\subsection*{User Prompt}

INPUTS:
\begin{itemize}
  \item A corpus of 30 Cell Painting-based mechanism assessment reports
  \item Each report typically includes: Mechanism verdict, feature-based evidence summary, a mechanistic linkage explanation and caveats or alternative hypotheses.
\end{itemize}

TASK: Evaluate the corpus for internal consistency across the following three dimensions. You are not judging scientific accuracy. Focus on mutual alignment in scientific reasoning, terminology, and structure.

\subsubsection*{1. Thematic Consistency}

Definition: Do all reports demonstrate same scientific purpose and consistent style of biological reasoning?

What to look for:
\begin{itemize}
  \item Reports clearly assess whether observed phenotypes support a hypothesized MoA
  \item Use of biological logic at cellular or subcellular level
  \item Use of structured reasoning patterns: “If MoA X is true, then we expect Y phenotype, which we observe.”;“This phenotype is consistent with known outcomes of X (e.g., ER stress)”
  \item Avoidance of vague or off-topic content
  \item Consistency in depth of biological interpretation
\end{itemize}

Scoring Guide:
\begin{itemize}
  \item \textbf{10}: Reports are clear, coherent, and mechanistic
  \item \textbf{8--9}: Mostly consistent with minor variation in depth
  \item \textbf{6--7}: Some reports are descriptive only
  \item \textbf{4--5}: Inconsistent styles or lack MoA focus
  \item \textbf{1--3}: Highly divergent in purpose or reasoning
\end{itemize}

\subsubsection*{2. Terminological Consistency}

Definition: Are technical terms, feature names, and mechanistic labels used consistently?

What to look for:
\begin{itemize}
  \item Uniform use of Cell Painting features (e.g., \texttt{Texture\_Entropy})
  \item Consistent naming of biological phenomena
  \item Standardized use of MoA terms (e.g., “oxidative stress”)
  \item Avoidance of ambiguous or informal language
  \item Quantitative descriptors (e.g., “+0.3 increase”) preferred over vague phrases
\end{itemize}

Scoring Guide:
\begin{itemize}
  \item 10: Terminology precise and consistent
  \item 8--9: Minor synonym use
  \item 6--7: Mixed naming for same terms
  \item 4--5: Frequent inconsistencies
  \item 1--3: Terms are vague or informal
\end{itemize}

\subsubsection*{3. Structural Consistency}

Definition: Do reports follow a similar structure and rhetorical flow?

What to look for:
\begin{itemize}
  \item Inclusion of key components: Mechanism verdict, evidence summary, mechanistic linkage, caveats or alternatives.
  \item Similar order, depth, and sentence structure
  \item Avoidance of missing or reordered sections
\end{itemize}

Scoring Guide:
\begin{itemize}
  \item \textbf{10}: Fully consistent structure
  \item \textbf{8--9}: Minor variation in order or depth
  \item \textbf{6--7}: Incomplete or reordered sections
  \item \textbf{4--5}: Frequent structural differences
  \item \textbf{1--3}: Highly inconsistent organization
\end{itemize}

\subsubsection*{Scoring Instructions}

\begin{itemize}
  \item Assign a score between \textbf{1--10} for each dimension
  \item Provide a concise justification (\textless 50 words)
  \item Return only the JSON in the following format
\end{itemize}

\begin{tcolorbox}[colback=gray!5, colframe=black!10!gray, title=JSON Output Format]
\begin{verbatim}
{
  "corpus_evaluation": {
    "Thematic Consistency": {
      "score": <1-10>,
      "justification": "<brief explanation>"
    },
    "Terminological Consistency": {
      "score": <1-10>,
      "justification": "<brief explanation>"
    },
    "Structural Consistency": {
      "score": <1-10>,
      "justification": "<brief explanation>"
    }
  }
}
\end{verbatim}
\end{tcolorbox}

\subsection*{Corpus Evaluation}
\begin{table}[H]
\centering
\caption{FeatRank Agent's Repeatability}
\label{FeatRank Agent's Repeatability}
\resizebox{\textwidth}{!}{%
\begin{tabular}{lcccccccc}
\toprule
\textbf{Temp 1} & \textbf{Features Number Avg} & \textbf{Features Number std}&  \textbf{Top 5 Stable Features}  \\
\midrule
0.0 & 18.37 & 2.37 & AreaShape\_Area, AreaShape\_Eccentricity, Texture\_Contrast\_5\_02\_256, Granularity\_2, Granularity\_3\\
0.0& 18.20 & 2.26 & AreaShape\_Area, AreaShape\_Eccentricity, Texture\_Contrast\_5\_02\_256, Granularity\_2, Granularity\_3 \\
0.0& 18.27 & 2.31 &AreaShape\_Area, AreaShape\_Eccentricity, Texture\_Contrast\_5\_02\_256, Granularity\_2, Granularity\_3 \\
0.0& 18.87 & 1.96 &AreaShape\_Area, AreaShape\_Eccentricity, Texture\_Contrast\_5\_02\_256, Granularity\_2, Granularity\_3 \\
0.0& 17.17 & 2.28 &AreaShape\_Area, AreaShape\_Eccentricity, Texture\_Contrast\_5\_02\_256, Granularity\_2, Granularity\_3 \\
\bottomrule
\end{tabular}
}
\end{table}
\begin{table}[H]
\centering
\caption{Corpus score comparison across different temperature setting.}
\resizebox{\textwidth}{!}{%
\label{tab:different_temperature}
\begin{tabular}{lcccccccc}
\toprule
\textbf{Temp1/Temp2} & \textbf{Thematic Consistency}  & \textbf{Terminological Consistency} & \textbf{Structural Consistency}&  \textbf{Averaged Corpus Score}  \\
\midrule
0.0/0.0 & 8.00 & 7.00 & 9.00 & 8.00 \\
0.0/0.1 & 8.00 & 7.00 & 8.00 & 7.67 \\
0.0/0.2 & 8.00 & 7.00 & 8.00 & 7.67 \\
0.0/0.5 & 8.00 & 7.00 & 8.00 & 7.67 \\
0.0/1.0 & 8.00 & 7.00 & 8.00 & 7.67 \\
\bottomrule
\end{tabular}
}
\end{table}

\section{Counterfactual Prompt Experiments: Dosage and Time}

To address the concern that the high retrieval performance may be due to metadata correlations (e.g., compound identity being indirectly inferred from time or dosage), rather than genuine multimodal alignment, we conducted controlled ablation experiments to isolate and evaluate the extent to which different textual components contribute to model performance. Specifically, we masked individual fields in the text prompts: compound name + MOA, concentration, or time—and measured the calculating the changes in retrieval accuracy.

As shown in Table~\ref{tab:mask_components}, masking the compound name and MOA (Experiment 1) results in a catastrophic performance drop (e.g., text-to-image R@1 drops from 98.70 to 3.50; MRR drops by nearly 90\%), indicating that the model relies heavily on compound-specific textual information. We mask the compound name and its mechanism of action (MOA) jointly, rather than separately, because these two fields are semantically correlated and often co-informative. This design choice is consistent with our setup in the drug classification experiment (Table ~\ref{classification-table}), and ensures a fair and aligned evaluation across experiments. In contrast, masking concentration or time results in only moderate to negligible performance degradation (e.g., R@1 drops to 3.50 when masking compound, and only to 93.00 when masking time). This pattern suggests that while the model encodes and leverages compound identity meaningfully. To further test CP-CLIP’s robustness to misleading contextual cues, we designed counterfactual prompt experiments in two settings:
\begin{itemize}
    \item \textbf{Clean Prompt:} The target classification field was masked, but all other contextual metadata (e.g., time, channel) remained correct.
    \item \textbf{Disturbed Prompt:} The target field was masked, and unrelated metadata fields were shuffled randomly, keeping only the compound identity correct. Since we have demonstrated that compound identity has a strong influence on retrieval performance.
\end{itemize}
\begin{table}[H]
\centering
\caption{Retrieval performance before and after masking different textual components. }
\small
\label{tab:mask_components}
\begin{tabular}{llcccc}
\toprule
\textbf{Experiment} & \textbf{Metric} & \textbf{Original} & \textbf{Masked} & \textbf{$\delta$ Absolute} & \textbf{$\delta$ Relative (\%)} \\
\midrule
\multicolumn{6}{l}{\textbf{Experiment 1: Mask Compound Name + MOA}} \\
\multirow{4}{*}{Text-to-Image}
 & R@1  & 98.70 & 3.50   & -95.20   & -96.45 \\
 & R@5  & 100.00 & 16.80  & -83.20   & -83.20 \\
 & R@10 & 100.00 & 29.60  & -70.40   & -70.40 \\
 & MRR  & 0.9935 & 0.1178 & -0.8757  & -88.15 \\
\multirow{4}{*}{Image-to-Text}
 & R@1  & 98.70 & 3.30   & -95.40   & -96.66 \\
 & R@5  & 100.00 & 16.10  & -83.90   & -83.90 \\
 & R@10 & 100.00 & 28.70  & -71.30   & -71.30 \\
 & MRR  & 0.9935 & 0.1102 & -0.8833  & -88.90 \\
 
\midrule
\multicolumn{6}{l}{\textbf{Experiment 2: Mask Concentration Only}} \\
\multirow{4}{*}{Text-to-Image}
 & R@1  & 98.70 & 57.00  & -41.70   & -42.25 \\
 & R@5  & 100.00 & 84.00  & -16.00   & -16.00 \\
 & R@10 & 100.00 & 91.40  & -8.60    & -8.60 \\
 & MRR  & 0.9935 & 0.6829 & -0.3106  & -31.26 \\
\multirow{4}{*}{Image-to-Text}
 & R@1  & 98.70 & 55.20  & -43.50   & -44.07 \\
 & R@5  & 100.00 & 79.20  & -20.80   & -20.80 \\
 & R@10 & 100.00 & 87.40  & -12.60   & -12.60 \\
 & MRR  & 0.9935 & 0.6617 & -0.3318  & -33.39 \\
 
\midrule
\multicolumn{6}{l}{\textbf{Experiment 3: Mask Time Only}} \\
\multirow{4}{*}{Text-to-Image}
 & R@1  & 98.70 & 93.00  & -5.70    & -5.78 \\
 & R@5  & 100.00 & 99.50  & -0.50    & -0.50 \\
 & R@10 & 100.00 & 100.00 & 0.00     & 0.00 \\
 & MRR  & 0.9935 & 0.9590 & -0.0345  & -3.47 \\
\multirow{4}{*}{Image-to-Text}
 & R@1  & 98.70 & 94.70  & -4.00    & -4.05 \\
 & R@5  & 100.00 & 100.00 & 0.00     & 0.00 \\
 & R@10 & 100.00 & 100.00 & 0.00     & 0.00 \\
 & MRR  & 0.9935 & 0.9726 & -0.0209  & -2.11 \\
 
\bottomrule
\end{tabular}
\end{table}

\begin{table}[H]
\caption{Concentration Classification under Masked and Counterfactual Prompts}
\label{sup:concentration_classification_full}
\centering
\resizebox{\textwidth}{!}{%
\begin{tabular}{lccccccc}
\toprule
Concentration & Firocoxib  & Opicapone & Cinoxacin & Neratinib & Hydroflumethiazide & Acetaminophen & Primidone \\
\midrule
0.00316µM  CLIP (disturb) & 0.4866 & 0.3341 & 0.3889 & 0.2965 & 0.1634 & 0.2298 & 0.2174 \\
\rowcolor{gray!10} 0.00316µM  CP-CLIP (disturb)& 0.4663 & 0.5052 & 0.4890 & 0.4887 & 0.2350 & 0.3155 & 0.3866 \\
0.00316µM  CLIP & 0.7371 & 0.5203 & 0.6491 & 0.5209 & 0.4780 & 0.4237 & 0.4167 \\
\rowcolor{gray!20} 0.00316µM  CP-CLIP  & 0.6866 & 0.7307 & 0.6211 & 0.6233 & 0.2819 & 0.3909 & 0.5349 \\

0.01µM  CLIP (disturb)& 0.5588 & 0.2708 & 0.1757 & 0.3731 & 0.2147 & 0.2161 & 0.2612 \\
\rowcolor{gray!10} 0.01µM  CP-CLIP (disturb)& 0.3212 & 0.4409 & 0.3694 & 0.4677 & 0.4906 & 0.4598 & 0.2321 \\
0.01µM  CLIP  & 0.7887 & 0.5391 & 0.3758 & 0.6267 & 0.4383 & 0.3473 & 0.5000 \\
\rowcolor{gray!20} 0.01µM  CP-CLIP  & 0.4356 & 0.6343 & 0.4913 & 0.5521 & 0.5783 & 0.5561 & 0.2635 \\

0.0316µM  CLIP (disturb)& 0.5714 & 0.1921 & 0.3178 & 0.2890 & 0.2017 & 0.2873 & 0.1670 \\
\rowcolor{gray!10} 0.0316µM  CP-CLIP (disturb)& 0.4335 & 0.4714 & 0.3928 & 0.7949 & 0.3725 & 0.5151 & 0.4015 \\
0.0316µM  CLIP  & 0.7437 & 0.4749 & 0.5839 & 0.5330 & 0.5033 & 0.5387 & 0.4628 \\
\rowcolor{gray!20} 0.0316µM  CP-CLIP  & 0.5754 & 0.6048 & 0.6374 & 0.9205 & 0.4771 & 0.6364 & 0.4773 \\

0.1µM  CLIP (disturb)& 0.2992 & 0.2652 & 0.3369 & 0.2866 & 0.2486 & 0.2232 & 0.1938 \\
\rowcolor{gray!10} 0.1µM  CP-CLIP (disturb)& 0.4660 & 0.2935 & 0.4015 & 0.3035 & 0.4051 & 0.3000 & 0.4148 \\
0.1µM  CLIP  & 0.4775 & 0.3554 & 0.3690 & 0.5203 & 0.2819 & 0.4823 & 0.2661 \\
\rowcolor{gray!20} 0.1µM  CP-CLIP  & 0.6402 & 0.4459 & 0.5442 & 0.4974 & 0.5014 & 0.3832 & 0.4971 \\

0.316µM  CLIP (disturb)& 0.3343 & 0.2301 & 0.2776 & 0.2131 & 0.2505 & 0.1994 & 0.2961 \\
\rowcolor{gray!10} 0.316µM  CP-CLIP (disturb)& 0.5079 & 0.3333 & 0.4027 & 0.4595 & 0.2848 & 0.4962 & 0.2642 \\
0.316µM  CLIP  & 0.5401 & 0.4179 & 0.6288 & 0.4591 & 0.6056 & 0.4077 & 0.5475 \\
\rowcolor{gray!20} 0.316µM  CP-CLIP  & 0.6272 & 0.4686 & 0.5068 & 0.6438 & 0.4399 & 0.5477 & 0.3217 \\

1.0µM  CLIP (disturb)& 0.4430 & 0.2748 & 0.2505 & 0.5455 & 0.1803 & 0.3082 & 0.3073 \\
\rowcolor{gray!10} 1.0µM  CP-CLIP (disturb)& 0.6364 & 0.4411 & 0.4103 & 0.4792 & 0.4459 & 0.3974 & 0.3702 \\
1.0µM  CLIP  & 0.6611 & 0.5773 & 0.6211 & 0.8856 & 0.4909 & 0.5766 & 0.4473 \\
\rowcolor{gray!20} 1.0µM  CP-CLIP  & 0.8137 & 0.6469 & 0.6500 & 0.6217 & 0.5133 & 0.4469 & 0.5673 \\

3.162µM  CLIP (disturb) & 0.3219 & 0.2933 & 0.2815 & 0.2119 & 0.2016 & 0.2635 & 0.2114 \\
\rowcolor{gray!10} 3.162µM  CP-CLIP (disturb) & 0.5483 & 0.3324 & 0.2900 & 0.4479 & 0.2600 & 0.1577 & 0.3991 \\
3.162µM  CLIP & 0.5528 & 0.5902 & 0.5459 & 0.6442 & 0.5826 & 0.5459 & 0.3255 \\
\rowcolor{gray!20} 3.162µM  CP-CLIP & 0.7599 & 0.5243 & 0.4577 & 0.5812 & 0.3433 & 0.2415 & 0.4454 \\

10.0µM  CLIP (disturb)& 0.3871 & 0.4191 & 0.2997 & 0.1502 & 0.1696 & 0.1813 & 0.2152 \\
\rowcolor{gray!10} 10.0µM  CP-CLIP (disturb)& 0.6340 & 0.3924 & 0.6029 & 0.5010 & 0.3401 & 0.3562 & 0.3499 \\
10.0µM  CLIP  & 0.7079 & 0.7125 & 0.5729 & 0.6616 & 0.3300 & 0.3685 & 0.5673 \\
\rowcolor{gray!20} 10.0µM  CP-CLIP  & 0.7895 & 0.6469 & 0.6500 & 0.5874 & 0.5133 & 0.4469 & 0.3886 \\
\bottomrule
\end{tabular}
}
\end{table}



\vspace{-2em}
\begin{table}[ht]
\centering
\caption{Time Comparison (clean vs disturb prompt): CLIP vs CP-CLIP}
\label{tab:time-comparison_full}
\scalebox{0.6}{ 
\begin{tabular}{lccccccc}
\toprule
Time & Ixabepilone  & Methoxsalen & Sulfinpyrazone & Triamterene & Miconazole & Ceritinib & Acetohexamide \\
\midrule
24h CLIP (disturb) & 0.7513 & 0.5903 & 0.6473 & 0.7212 & 0.6882 & 0.6319 & 0.7650 \\
\rowcolor{gray!20} 24h CP-CLIP (disturb) & 0.9600 & 0.9556 & 0.9101 & 0.8844 & 0.9309 & 0.9323 & 0.9309 \\
24h CLIP & 0.9950 & 1.0000 & 1.0000 & 1.0000 & 1.0000 & 0.9950 & 1.0000 \\
\rowcolor{gray!20} 24h CP-CLIP & 0.9980 & 1.0000 & 1.0000 & 1.0000 & 1.0000 & 1.0000 & 1.0000 \\
48h CLIP (disturb) & 0.7586 & 0.6829 & 0.6218 & 0.6979 & 0.6322 & 0.6927 & 0.7650 \\
\rowcolor{gray!20} 48h CP-CLIP (disturb) & 0.9600 & 0.9592 & 0.9238 & 0.8856 & 0.9340 & 0.9375 & 0.9387 \\
48h CLIP & 0.9950 & 1.0000 & 1.0000 & 1.0000 & 1.0000 & 0.9950 & 1.0000 \\
\rowcolor{gray!20} 48h CP-CLIP & 0.9980 & 1.0000 & 1.0000 & 1.0000 & 1.0000 & 1.0000 & 1.0000 \\
\bottomrule
\end{tabular}
}
\end{table}

\begin{table}[ht]
\centering
\caption{Summary of Concentration Classification Performance: CP-CLIP vs CLIP under clean vs. disturbed prompts.}
\label{tab:summary-comparison_conc}
\small
\scalebox{0.9}{ 
\begin{tabular}{l|ccc|ccc}
\toprule
\textbf{Compound} & 
\multicolumn{3}{c|}{\textbf{CLIP}} & 
\multicolumn{3}{c}{\textbf{CP-CLIP}} \\
 & Clean & Disturbed & $\Delta$ & Clean & Disturbed & $\Delta$ \\
\midrule
firocoxib           & 0.5637 & 0.4163 & -0.1474 & 0.6025 & 0.5038 & -0.0987 \\
opicapone           & 0.4244 & 0.2831 & -0.1413 & 0.4631 & 0.3956 & -0.0675 \\
cinoxacin           & 0.3962 & 0.2875 & -0.1087 & 0.4500 & 0.4231 & -0.0269 \\
neratinib           & 0.5131 & 0.2875 & -0.2256 & 0.5269 & 0.4950 & -0.0319 \\
hydroflumethiazide  & 0.3669 & 0.2025 & -0.1644 & 0.3350 & 0.3650 & +0.0300 \\
acetaminophen       & 0.3531 & 0.2381 & -0.1150 & 0.3762 & 0.3844 & +0.0082 \\
primidone           & 0.3550 & 0.2350 & -0.1200 & 0.3588 & 0.3556 & -0.0032 \\
\midrule
CP-CLIP $\uparrow$ over CLIP (Clean)     & \multicolumn{3}{c|}{+7.83\% Accuracy} & \multicolumn{3}{c}{+7.72\% F1-Score} \\
CP-CLIP $\uparrow$ over CLIP (Disturbed) & \multicolumn{3}{c|}{+54.05\% Accuracy} & \multicolumn{3}{c}{+50.55\% F1-Score} \\
\bottomrule
\end{tabular}%
}
\end{table}

\begin{table}[H]
\centering
\caption{Summary of Time Classification Performance: CP-CLIP vs CLIP under clean vs. disturbed prompts.}
\label{tab:summary-comparison_time}
\small
\scalebox{0.9}{ 
\begin{tabular}{l|ccc|ccc}
\toprule
\textbf{Compound} & 
\multicolumn{3}{c|}{\textbf{CLIP}} & 
\multicolumn{3}{c}{\textbf{CP-CLIP}} \\
 & Clean & Disturbed & $\Delta$ & Clean & Disturbed & $\Delta$ \\
\midrule
ixabepilone       & 0.9950 & 0.7550 & -0.2400 & 0.9975 & 0.9600 & -0.0375 \\
methoxsalen       & 1.0000 & 0.6425 & -0.3575 & 1.0000 & 0.9575 & -0.0425 \\
sulfinpyrazone    & 1.0000 & 0.6350 & -0.3650 & 1.0000 & 0.9175 & -0.0825 \\
triamterene       & 1.0000 & 0.7100 & -0.2900 & 1.0000 & 0.8850 & -0.1150 \\
miconazole        & 1.0000 & 0.6625 & -0.3375 & 1.0000 & 0.9325 & -0.0675 \\
ceritinib         & 0.9950 & 0.6650 & -0.3300 & 1.0000 & 0.9350 & -0.0650 \\
acetohexamide     & 1.0000 & 0.7650 & -0.2350 & 1.0000 & 0.9350 & -0.0650 \\
\midrule
CP-CLIP $\uparrow$ over CLIP (Clean)     & \multicolumn{3}{c|}{+0.08\% Accuracy} & \multicolumn{3}{c}{+0.08\% F1-Score} \\
CP-CLIP $\uparrow$ over CLIP (Disturbed) & \multicolumn{3}{c|}{+34.91\% Accuracy} & \multicolumn{3}{c}{+35.23\% F1-Score} \\
\bottomrule
\end{tabular}%
}
\end{table}

This setup isolates the model’s reliance on different types of metadata and tests whether it can still make accurate predictions under misleading or noisy context. For the concentration classification task (Table~\ref{tab:summary-comparison_conc}), CP-CLIP achieves a +7.72\% F1-score improvement over CLIP under clean prompts, and a +50.55\% F1-score gain under disturbed prompts. Similarly, for the time classification task (Table~\ref{tab:summary-comparison_time}), CP-CLIP maintains stable performance with only marginal degradation under disturbed prompts, achieving a +35.23\% F1-score improvement over CLIP. The detailed scores for each category are available in Table~\ref{sup:concentration_classification_full} and Table~\ref{tab:time-comparison_full}. In contrast, CLIP exhibits substantial performance drops in disturbed scenarios, suggesting that it is more susceptible to spurious correlations in the metadata. 

This results confirm that CP-CLIP is not learning shortcuts based on metadata correlations but is instead capturing robust multimodal associations between visual morphology and semantic input prompts. The counterfactual experiment and contextual masking experiments effectively disentangles various sources of information, demonstrating that CP-CLIP can generalize to various classification tasks even under adversarial or misleading contextual conditions.

\section{Retrieval Performance Benchmarks}

In addition to the classification accuracy metrics presented in Table~\ref{classification-table} and Table~\ref{unseen-drug}, which are used for benchmarking in-distribution performance and out-of-distribution performance across various conditions (e.g., cell line, channel, and compound), we further report retrieval-based metrics to evaluate the effectiveness of contrastive learning models.

Specifically, we use Recall@K and Mean Reciprocal Rank (MRR) on a held-out validation set to assess both text-to-image (T→I) and image-to-text (I→T) retrieval performance. These metrics provide a complementary perspective on model alignment quality between visual and textual modalities, particularly in settings where ranking-based retrieval is desirable.

Table summarizes the retrieval performance of several contrastive models, including CLIP, SigLIP, and variants of our proposed CP-CLIP, under both text-to-image and image-to-text settings (Table~\ref{tab:recallK}). To provide a more complete picture, we also include the retrieval performance of CLOOME, shown in Table~\ref{tab:cloome}. While CLOOME is not a general-purpose contrastive model, we report its performance on molecule–image and image–molecule retrieval tasks for completeness. Notably, CLOOME’s retrieval is limited to molecular inputs only, and is not applicable for cell-line or other biological metadata queries.

\begin{table}[H]
\centering
\caption{Performance of CLOOME on Molecule–Image and Image–Molecule Retrieval Tasks.}
\resizebox{\textwidth}{!}{%
\label{tab:cloome}
\begin{tabular}{lcccccc|cccccc}
\toprule
\textbf{Model} & \multicolumn{6}{c|}{\textbf{Molecule-Image}} & \multicolumn{6}{c}{\textbf{Image-Molecule}} \\
\cmidrule(lr){2-7} \cmidrule(lr){8-13}
& R@1 & R@5 & R@10 & R@20 & R@50 & MRR & R@1 & R@5 & R@10 & R@20 & R@50 & MRR \\
\midrule
CLOOME & 95.58 & 99.94 & 99.94 & 99.94 & 99.94 & 0.9754 & 59.86  & 59.98 & 60.52 & 65.59 & 71.89 & 0.6063   \\
\bottomrule
\end{tabular}
}
\label{tab:wide_retrieval}
\end{table}

\begin{table}[H]
\centering
\caption{Context-to-Image (T-I) and Image-to-Context (I-T) retrieval performance using Recall@K.}
\resizebox{\textwidth}{!}{%
\label{tab:recallK}
\begin{tabular}{lcccccc|cccccc}
\toprule
\textbf{Model} & \multicolumn{6}{c|}{\textbf{Context-Image}} & \multicolumn{6}{c}{\textbf{Image-Context}} \\
\cmidrule(lr){2-7} \cmidrule(lr){8-13}
& R@1 & R@5 & R@10 & R@20 & R@50 & MRR & R@1 & R@5 & R@10 & R@20 & R@50 & MRR \\
\midrule
CLIP ViT-B/16 & 66.8 & 90.80 & 95.40 & 97.89 & 99.38 & 0.7719  & 58.55 & 80.67 & 86.54 & 91.71 & 96.85 & 0.6820  \\
SigLIP-ViT-B/16 & 43.85 & 67.38 & 77.44 & 85.75 & 93.19 & 0.5457 & 38.65 & 56.15 & 63.86 & 71.56 & 81.92 & 0.4700 \\
SigLIP-ViT-B/16 (D) & 25.93 & 52.21 & 61.37 & 71.45 & 85.38 & 0.3842 & 20.71 & 39.87 & 48.83 & 59.65 & 74.33 & 0.3015  \\
CP-CLIP ViT-B/16 (fp) & 72.97 & 93.96 & 97.47 & 99.10 & 99.86 & 0.8213 & 64.20 & 86.55 & 91.99 & 95.73 & 98.75 & 0.7385\\
CP-CLIP ViT-B/16 (D) & 77.09 & 94.69 & 97.87 & 99.21 & 99.74 & 0.8479  & 68.92 & 87.77 & 92.14 & 95.56 & 98.55& 0.7716 \\
CP-CLIP ViT-L/16 (D) & 73.83 & 92.93 & 96.44 & 98.49 & 99.61 & 0.8215 & 64.77 & 85.02 & 90.13 & 93.83 & 97.52 & 0.7351 \\
\bottomrule
\end{tabular}
}
\label{tab:wide_retrieval}
\end{table}


\end{document}